\documentclass[sigconf]{acmart}
\usepackage{multicol}
\usepackage{multirow}
\usepackage{color}
\usepackage{algorithm}
\usepackage{algorithmic}
\usepackage{caption}
\usepackage{subcaption}

\definecolor{morange}{RGB}{253,146,38}
\definecolor{mgreen}{RGB}{26,175,84}

\newcommand{\haojie}[1]{{\color{black}#1}}

\AtBeginDocument{%
  \providecommand\BibTeX{{%
    \normalfont B\kern-0.5em{\scshape i\kern-0.25em b}\kern-0.8em\TeX}}}

\copyrightyear{2021}
\acmYear{2021}
\setcopyright{acmcopyright}
\acmConference[CIKM '21] {Proceedings of the 30th ACM International Conference on Information and Knowledge Management}{November 1--5, 2021}{Virtual Event, Australia.}
\acmBooktitle{Proceedings of the 30th ACM Int'l Conf. on Information and Knowledge Management (CIKM '21), November 1--5, 2021, Virtual Event, Australia}
\acmPrice{15.00}
\acmDOI{10.1145/3459637.3481932}
\acmISBN{978-1-4503-8446-9/21/11}

\begin{document}
\fancyhead{}

\title{Learning to Expand: Reinforced Response Expansion for Information-seeking Conversations}

\newcommand\blfootnote[1]{%
  \begingroup
  \renewcommand\thefootnote{}\footnote{#1}%
  \addtocounter{footnote}{-1}%
  \endgroup
}

\author{Haojie Pan$^1$, Cen Chen$^{2\dagger}$, Chengyu Wang$^1$, Minghui Qiu$^{1}$, Liu Yang$^3$, Feng Ji$^1$, Jun Huang$^1$}
\affiliation{$^1$ Alibaba Group $^2$ East China Normal
University, China $^3$ University of Massachusetts at Amherst
\institution{\{haojie.phj,chengyu.wcy,minghui.qmh,zhongxiu.jf,huangjun.hj\}@alibaba-inc.com}
\city{cenchen@dase.ecnu.edu.cn, yangliuyx@gmail.com}
\country{~}
}

\renewcommand{\shortauthors}{H. Pan et al.}
\renewcommand{\authors}{H. Pan, C. Chen, C. Wang, M. Qiu, L. Yang, F. Ji, J. Huang}

\begin{abstract}
Information-seeking conversation systems are increasingly popular in real-world applications, especially for e-commerce companies. 
To retrieve appropriate responses for users, it is necessary to compute the matching degrees between candidate responses and users' queries with historical dialogue utterances. 
As the contexts are usually much longer than responses, it is thus necessary to expand the responses (usually short) with richer information. Recent studies on pseudo-relevance feedback (PRF) have demonstrated its effectiveness in query expansion for search engines, hence we consider expanding response using PRF information.
However, existing PRF approaches are either based on heuristic rules or require heavy manual labeling, which are not suitable for solving our task.
To alleviate this problem, we treat the PRF selection for response expansion as a learning task and propose a reinforced learning method that can be trained in an end-to-end manner without any human annotations. 
More specifically, we propose a reinforced selector to extract useful PRF terms to enhance response candidates and a BERT-based response ranker to rank the PRF-enhanced responses. The performance of the ranker serves as a reward to guide the selector to extract useful PRF terms, which boosts the overall task performance.
Extensive experiments on both standard benchmarks and commercial datasets prove the superiority of our reinforced PRF term selector compared with other potential soft or hard selection methods. 
Both case studies and quantitative analysis show that our model is capable of selecting meaningful PRF terms to expand response candidates and also achieving the best results compared with all baselines on a variety of evaluation metrics. We have also deployed our method on online production in an e-commerce company, which shows a significant improvement over the existing online ranking system.\blfootnote{$^{\dagger}$~Cen Chen is the corresponding author.}
\end{abstract}



\maketitle

\section{Introduction}

\begin{figure}[t]
\centering
\includegraphics[width=0.9\linewidth]{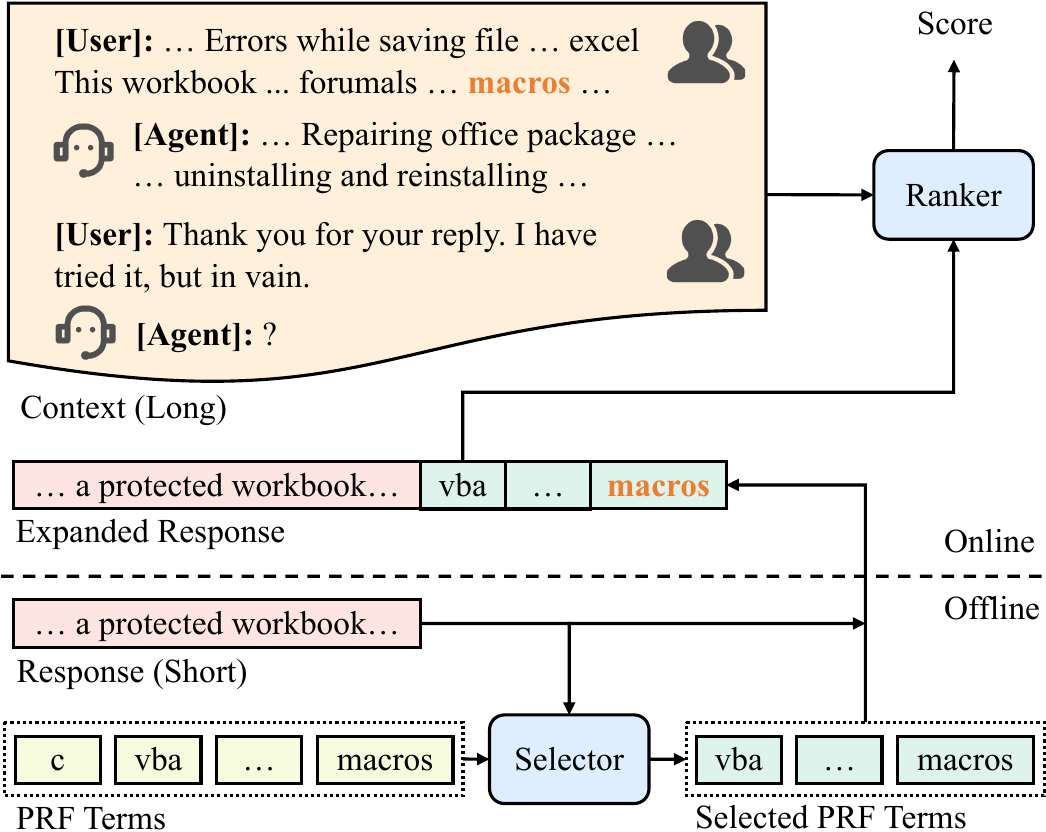}
\caption{Overview of our proposed model ``PRF-RL''. 
}
\label{fig:overview}
\vspace{-.5em}
\end{figure}

Intelligent personal assistant systems or chatbots such as Microsoft's XiaoIce, Apple's Siri, Google's Assistant have boomed during the last few years. It brings the interests from both academia and industry on information-seeking conversational systems, where end-users can access information with conversational interactions with the systems. 
Most of the assistant systems are based on retrieval-based methods as they can produce more informative, relevant, fluent, and controllable responses~\cite{alime-demo}.
The common practice for these methods is to use historical dialogue utterances and the current user query as the ``context'' and use it to match the most relevant ``response'' from the candidates in databases~\cite{DBLP:conf/sigir/YangQQGZCHC18,DBLP:conf/www/0005QQCGZCC20}. 

Recently, researchers have shown that external Pseudo-relevance Feedback (PRF) expansion is beneficial for context-response matching in multi-turn context understanding~\cite{DBLP:conf/sigir/YangQQGZCHC18}. The goal of PRF expansion is to extract PRF terms from relevant documents to improve the original ``query'' representations. 
However, we find that it is also an important task to expand the ``response'' instead of the ``query'' for info-seeking conversations. The reasons are two-fold.
First, the context is usually much longer than the response, which leads to a vocabulary gap between the context and the response. For example, as shown in Figure \ref{fig:overview}, some words in the context contain key information, such as the term ``macros'', which is relevant to ``protected workbook'' in the response but is not explicitly shown. Hence, it is important to expand the responses with richer information to better match the contexts. 
Second, 
to expand response is more efficient than query expansion for online retrieval systems as it requires no additional process of the input query, which makes it more suitable for real-world applications.

Despite the importance of response expansion for info-seeking conversations, few studies address this problem. To bridge this gap, this study seeks to examine the helpfulness of response expansion with external PRF information. The task is a non-trivial task that mainly has two challenges. First, existing studies~\cite{DBLP:conf/sigir/YangQQGZCHC18} for PRF terms expansion are not directly applicable for the task, as these studies do not select high-quality PRF terms and simply use all the PRF results as additional inputs. As shown in~\cite{Cao:2008:SGE:1390334.1390377}, some retrieved PRF terms are \textit{unrelated} to the query, thus not helpful for the retrieval system. Secondly, after high-quality PRF terms are produced, it is challenging to model the interactions between the context, the response, and the PRF terms. Furthermore, it is a labor-intensive task to manually examine which PRF terms are suitable for a given response. There is a compelling need to automatically extract helpful PRF terms for response expansion without labeled signals. 


To mitigate these challenges, we consider the response expansion problem as a learning task and build a effective way to guide the selection without explicitly labeled signals. The overview of our framework is outlined in Figure \ref{fig:overview}, which contains a PRF term selector for response expansion and a ranker for response ranking. 
Firstly, we consider Reinforcement Learning (RL) for our task, due to its effectiveness in exploration with implicit feedback~\cite{DBLP:journals/corr/abs-1708-02383,DBLP:journals/corr/FanTQBL17,DBLP:conf/aaai/WangYGWKZCTZJ18}. 
We model the problem as a Markov Decision Process (MDP) and propose a general framework to jointly learn a \textit{reinforced selector} with a \textit{response ranker}. We treat the reinforced PRF term selector as the agent that takes the actions to select a subset of PRF terms based on the state representation. These selected PRF terms will be further fed into the ranking module together with the conversation context and the response. The ranking module and the validation data play a role of an environment and output rewards for those selection actions.  The resulting rewards can guide the reinforced selector to generate higher-quality PRF terms to improve the ranking results. 
Secondly, we adopt BERT~\cite{DBLP:conf/naacl/DevlinCLT19} with different input formats as the encoder \haojie{of} our response ranker, since the language models pre-trained on large unsupervised corpora ~\cite{radford2019language,DBLP:journals/corr/abs-1907-11692} has demonstrated stunning success in NLP. 
After training, the selector can generate the response expansion offline and save them in databases, hence the efficiency of the online ranker will not be affected. We refer to our proposed whole model as ``\textbf{PRF-RL}''. 

We conduct extensive experiments on both public and industrial datasets, i.e., the MSDialog dataset  ~\cite{DBLP:conf/sigir/QuYCTZQ18} that contains customer service dialogs from Microsoft Answers community and the AliMe-CQA dataset ~\cite{alime-demo} collected from the chat logs between the customers and the service assistant agent in an e-commerce APP. We compare our method with various baselines, such as traditional retrieval models, neural ranking models, a strong multi-turn conversation response ranking baseline ~\cite{DBLP:conf/acl/WuLCZDYZL18}
and pre-trained language model based methods. Our method outperforms all the baseline methods on a variety of evaluation metrics. Besides, to demonstrate the effectiveness of our reinforced PRF term selector, we also compare our model with other potential PRF selection methods, including the rule-based selection method mentioned in ~\cite{DBLP:conf/sigir/YangQQGZCHC18}, soft and hard selection by gate functions such as \texttt{tanh} or Gumbel softmax trick~\cite{DBLP:conf/iclr/JangGP17}.
Both the numerical results and case studies show the superiority of our proposed reinforced selector.
We have deployed our proposed "\textbf{PRF-RL}" model in an online information-seeking system on a real e-commerce production AliMe. The online A/B test results show that our model significantly outperforms the existing online ranking system.

In a nutshell, our contributions can be summarized as follows:

(1) Our study is one of the earliest attempts to analyze how response expansion can be better utilized to improve BERT-based retrieval models. 
Response expansion is more efficient for online retrieval systems than query expansion, as it requires no additional process of the input query, which makes it more suitable for building real-world applications.
    
(2) We tackle the response expansion problem from a learning perspective and propose a novel method, i.e., PRF-RL, for the problem. Our method consists of two modules: i) a reinforced selector to extract useful PRF terms, and ii) a BERT response ranker with PRF. To the best of our knowledge, it is the first work that employs an RL-based strategy to select high-quality PRF terms for response ranking in the information-seeking systems.

(3) Experimental results on both public and industrial datasets show that our methods outperform various baselines and show that our reinforced PRF terms selector is superior to other competing PRF term selection mechanisms. We have also deployed our method on an online chatbot system in an e-commerce company.~\footnote{The source code of our method will be released in EasyTransfer~\cite{easytransfer}.} 
\section{Related Work}
\label{sec:related-work}

We summarize the related works on conversational information-seeking systems, query expansion, reinforcement learning and neural response ranking.

\subsection{Conversational Information-seeking Systems and Query Expansion}

Our research is relevant to conversational information-seeking systems.
\citet{radlinski2017theoretical} described the basic features of conversational information-seeking systems.  \citet{thomas2017misc} released the Microsoft Information-Seeking Conversation (MISC) dataset, which contains information-seeking conversations with a human intermediary. \citet{zhang2018towards} introduced the System Ask User Respond (SAUR) paradigm for conversational search and recommendation. In addition to conversational search models, researchers have also studied the medium of conversational search. \citet{spina2017extracting} studied the ways of presenting search results over speech-only channels to support conversational search ~\cite{trippas2015towards,spina2017extracting}. 

In conversational information-seeking systems, the context can be longer than the candidate responses. Hence, it is necessary to expand the responses. In IR systems, pseudo-relevance feedback (PRF) has been demonstrated the effectiveness for query expansion ~\cite{Cao:2008:SGE:1390334.1390377,Lavrenko:2001:RBL:383952.383972,Lv:2009:CSM:1645953.1646259, rocchio71relevance, Zamani:2016:PFB:2983323.2983844, Zhai:2001:MFL:502585.502654}. \citet{Cao:2008:SGE:1390334.1390377} incorporates multiple manual features to identify useful expanding terms. \citet{Lv:2009:CSM:1645953.1646259} compares methods for estimating query language models with pseudo-relevance feedback in ad-hoc information retrieval. 
\citet{DBLP:conf/emnlp/LiSHWHYSX18} proposed an end-to-end neural model to make the query directly interacts with the retrieved documents. 
The idea of query expansion using PRF inspires us to use the candidate response as a query to retrieve relevant terms, thus boosting ranking performance by reducing the problem of vocabulary mismatch between the context and the original response candidates.




\subsection{Reinforcement Learning}
Reinforcement Learning (RL) is a series of goal-oriented algorithms that have been studied for many decades in many disciplines ~\cite{DBLP:journals/spm/ArulkumaranDBB17}. The recent development in deep learning has greatly contributed to this area and has delivered amazing achievements in many domains, such as playing games against humans ~\cite{silver2017mastering}. There are two lines of work in RL: value-based methods and policy-based methods. Value-based methods, including SARSA ~\cite{rummery:cuedtr94} and the Deep Q Network ~\cite{DBLP:journals/nature/MnihKSRVBGRFOPB15}, take actions based on estimations of expected long-term return. On the other hand, policy-based methods such as REINFORCE ~\cite{DBLP:journals/ml/Williams92} optimize for a strategy that can map states to actions that promise the highest reward. 
It is proved that reinforcement learning is effective in data selection problems over many areas, such as active learning ~\cite{DBLP:journals/corr/abs-1708-02383}, co-training ~\cite{DBLP:conf/naacl/WuLW18}, and other applications of supervised learning~\cite{DBLP:conf/aaai/WangYGWKZCTZJ18,DBLP:conf/aaai/FengHZYZ18}.
Our proposed reinforced PRF term selector is trained by REINFORCE.

\subsection{Neural Response Ranking}
There is growing interest in research about conversation response generation and ranking with deep learning and reinforcement learning ~\cite{DBLP:journals/corr/abs-1809-08267}. There are two main categories of the previous works, including retrieval-based methods ~\cite{Tao:2019:MFN:3289600.3290985, DBLP:conf/acl/WuWXZL17, DBLP:conf/sigir/YanSW16, DBLP:conf/sigir/YanZE17, DBLP:conf/sigir/YangQQGZCHC18, DBLP:conf/emnlp/ZhouDWZYTLY16} and generation-based methods \cite{alime-chat, DBLP:conf/emnlp/RitterCD11, DBLP:conf/acl/ShangLL15,DBLP:journals/corr/VinyalsL15}. Our research work is related to retrieval-based methods. There has been some research on response ranking in multi-turn conversations with retrieval-based methods. \citet{DBLP:conf/sigir/YangQQGZCHC18} studied how to integrate external knowledge into deep neural networks for response ranking in information-seeking conversations. ~\cite{DBLP:conf/acl/WuLCZDYZL18} investigated matching a response with conversation contexts with dependency information learned by attention mechanisms of Transformers. The model proposed in this paper incorporating a reinforced PRF terms selection mechanism to select meaningful PRF terms to boost the performance of pre-trained model based response ranking model.
Recently, language models pre-trained on massive unsupervised corpora ~\cite{Peters:2018, DBLP:conf/naacl/DevlinCLT19, radford2019language, DBLP:journals/corr/abs-1907-11692,DBLP:conf/nips/YangDYCSL19,DBLP:journals/corr/abs-1909-11942} has achieved a significant improvement in many natural language processing tasks, ranging from syntactic parsing to natural language inference ~\cite{Peters:2018, DBLP:conf/naacl/DevlinCLT19}, as well as machine reading comprehension~\cite{DBLP:conf/naacl/DevlinCLT19, DBLP:conf/naacl/XuLSY19}, information retrieval tasks ~\cite{passrerankbert19, SimpleAppBERTDocRetreive19}. The pre-trained models such as BERT ~\cite{DBLP:conf/naacl/DevlinCLT19} are also applied to response selection ~\cite{JesseVig19,DBLP:journals/corr/abs-1908-04812,DBLP:conf/acl/HendersonVGCBCS19}. Our model applies BERT as a part of the response ranker with the response expanded by selected PRF terms. 
\section{Our Approach} \label{sec-approach}
\label{sec:model}
In this section, we formally present our PRF-RL model. We begin with a brief problem definition, followed by the proposed method.

\subsection{Problem Definition}
We formulate the problem of response ranking with pseudo-relevance feedback as follows. Given an information-seeking conversation dataset $\mathcal{D} = \{(\mathcal{U}_i, r_i, y_i) \}_{i=1}^N$, where $\mathcal{U} = \{u_1, u_2, ..., u_m \}$ is a \haojie{$m$-turn} dialog context,  $u_i = \{w_{i, 1}^{(u)}, w_{i, 2}^{(u)}, ..., w_{i, L_{u_i}}^{(u)} \}$ as the utterance in the $i$-th turn of this dialog \haojie{and $L_{u_i}$ is the number of sub-tokens of this utterance}. $r = \{w_{1}^{(r)}, w_{2}^{(r)}, ..., w_{n}^{(r)} \}$ is a response candidate and $y \in \{0, 1\}$ is the corresponding label.  When the psudo-relevance feedback information is incorporated, for each response $r$, we have PRF term set $\mathcal{P} = \{p_1, p_2, ..., p_k\}$ and $p_i = \{w_{i, 1}^{(p)}, w_{i, 2}^{(p)}, ..., w_{i, L_{p_i}}^{(p)} \}$ is $i$-th PRF term and \haojie{and $L_{p_i}$ is the number of sub-tokens of this PRF term}. The task is to learn a model which has two sub-modules: (1) a selection module $f(\cdot)$ to select a meaningful subset $\mathcal{P}' \subseteq \mathcal{P}$ of PRF terms and (2) a ranking module $g(\cdot)$ with $\mathcal{D}$ and $\mathcal{P}'$ as inputs to rank the responses. Given $\mathcal{U}$ and $\mathcal{P}$, the model should be able to generate 
a prediction $\hat{y}$ for response $r$ for ranking with other candidate responses.

\begin{figure*}[t]
\centering
\includegraphics[width=1.0\linewidth]{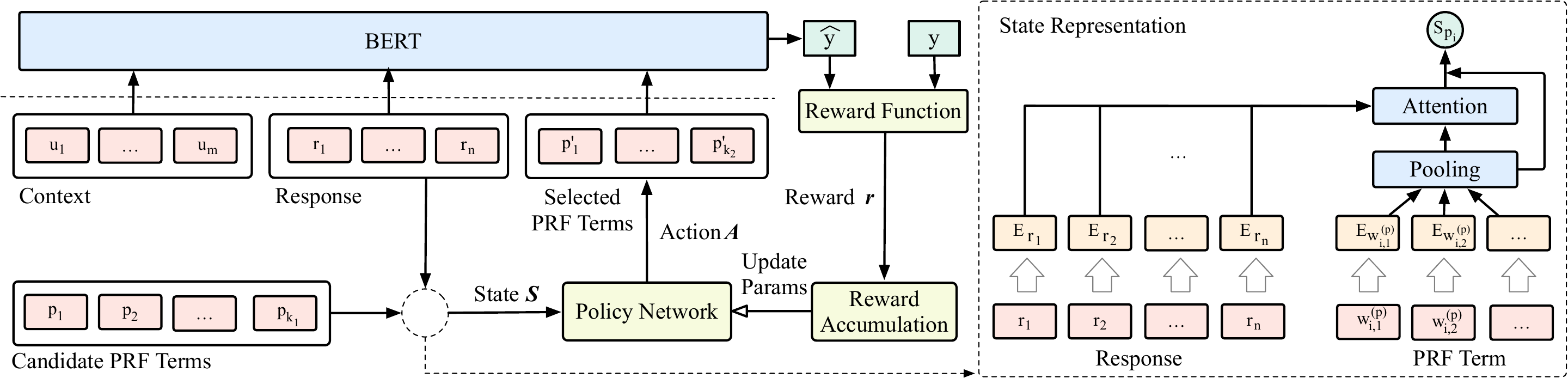}
\caption{Reinforced pseudo-relevance feedback selector.}
\label{fig:selector}
\end{figure*}


\subsection{Model Overview}
In the following sections, we describe our proposed model for response ranking with pseudo-relevance feedback. Given a set of retrieved PRF term candidates~\footnote{The details of how the candidates are retrieved and examples are shown in the experiment section.}, we first propose a reinforced PRF term selector to select a subset of the PRF terms, 
and feed them into a BERT-based 
response ranking model, which outputs the final predictions. 
We will first briefly introduce how the BERT response ranker works and then \haojie{leave more space to introduce our proposed reinforced PRF term selector in detail.} 
Figure \ref{fig:selector} presents an overview of the proposed method.

\subsection{BERT Response Ranker}
As mentioned above, once we have a context $\mathcal{U}$, a response $r$ and a set of PRF terms $\mathcal{P}$, we can concatenate them with the following specific format as the input of BERT $\mathbf{x} = \{$ \texttt{[CLS]}, $u_1$, \texttt{[EOT]}, $u_2$, \texttt{[EOT]}, ..., $u_m$, \texttt{[SEP]}, $r$, \texttt{[SEP]}, $p_1$, \texttt{[SEP]} $p_2$, \texttt{[SEP]}, ..., $p_k \}$. Here we use \texttt{[EOT]} to separate the multiple turns of the dialog and use \texttt{[SEP]} to separate different PRF terms since they are independent with each other. 
After multiple standard BERT layers, we can get a contextual representation $T_\texttt{[CLS]}$ of \texttt{[CLS]} token. We then feed the contextual representation into an extra feed forward network with sigmoid activation function to predict the ranking score of $r$ given $\mathcal{U}$ and $\mathcal{P}$, illustrated as follows:
\begin{align}
T_\texttt{[CLS]} &= \texttt{BERT}(\mathbf{x}). \\
\hat{y} = g(\mathcal{U}, \mathcal{P}, r) &= \sigma (\mathbf{W}_o^T T_\texttt{[CLS]} + \mathbf{b}).
\end{align}

The response ranking model can be optimized by gradient descent based methods and the cross-entropy loss is applied as follows.

\begin{equation}
Loss = - \left(
\sum_{i=1}^N y_i log (\hat{y}_i) + (1 - y_i) log (1 -  \hat{y}_i)
\right)
\label{eq:xent_loss}
\end{equation}

\subsection{Reinforced PRF Term Selector}

The PRF term selection process can be viewed as a sequential decision-making problem and modeled as a Markov Decision Process, which can be further solved by reinforcement learning. 
Under the reinforcement setting, the PRF term selector serves as the \textit{agent} and interacts with the environment consisting of the BERT response ranker and a utility evaluation dataset. With the help of a learnable policy network, the agent takes the \textit{actions} of selecting or dropping a given PRF term, where the decision is based on a \textit{state representation} of each candidate term. 
Then the BERT response ranker takes the selected PRF terms as one of the inputs and further provides \textit{rewards} for guiding the agent. The overview of the reinforced PRF term selector is shown in Figure \ref{fig:selector}.

To be specific, we formulate the learning framework as follows. Given a response $r$ and the corresponding PRF terms set $\mathcal{P} = \{ p_i \}_{i=1}^{k_1}$, where $k_1$ is the number of PRF terms before selection. We can apply a state representation network $z(r, p_i)$ for each PRF term $p_i$ to obtain states $\mathcal{S} = \{\mathbf{s}_i \}_{i=1}^{k_1}$ and feed each $\mathbf{s}_i$ into the policy network $\pi (\mathbf{s}_i) $. 
According to the policy, the agent takes the corresponding actions $\mathcal{A} = \{a_1, a_2, ..., a_{k_1} \}$, where $a_i \in \{0, 1\}$. 
The selected subset of PRF terms $\mathcal{P}' = \{ p_i | \forall p_i \in \mathcal{P}, a_i = 1 \}$, together with $\mathcal{U}$ and $r$, are then fed into the response ranker to output a prediction. The parameters of the response ranker will be updated with label $y$ and a reward $\mathbf{r}$ will be produced after evaluating the ranker's performance on a held-out validation dataset.

\subsubsection{State representation}
The state $\mathbf{s}_i = z(r, p_i)$ of given response $r$ and a PRF term $p_i$ is a $d$-dimension continuous real valued vector. We use the attention mechanism between token embeddings of $r$ and $p_i$ to obtain a contextual vector as follows.

With the WordPiece embeddings ~\cite{DBLP:journals/corr/WuSCLNMKCGMKSJL16}, we can represent tokens of $r$ and $p_i$ as:
\begin{align}
\mathbf{E_r} &= \{\mathbf{e}_{w_1^{(r)}}, \mathbf{e}_{w_2^{(r)}}, ..., \mathbf{e}_{w_n^{(r)}} \}, \\
\mathbf{E_{p_i}} &= \{\mathbf{e}_{w_{i, 1}^{(p)}}, \mathbf{e}_{w_{i, 2}^{(p)}}, ..., \mathbf{e}_{w_{i, L_{p_k}}^{(p)}} \}.
\end{align}
We first use pooling methods, e.g., max-pooling, to get a $d$-dimension vector $\mathbf{h_i} = pooling(\mathbf{E_{p_i}})$ to represent the PRF term and obtain the attended weighted context as follows:

\begin{equation}
\mathbf{h}_i' = \sum_{j=1}^{n} \frac{exp(\mathbf{h}_i \cdot \mathbf{e}_{w_j^{(r)}})}{\sum_{k=1}^nexp(\mathbf{h}_i \cdot \mathbf{e}_{w_k^{(r)}})} \mathbf{e}_{w_j^{(r)}}.
\end{equation}
Finally we get the state representation as $\mathbf{s}_i = \mathbf{h}_i + \mathbf{h}_i'$.

\subsubsection{Policy Network and Actions}
The agent (i.e., the reinforced selector) takes actions to decide whether to select the PRF term $p_i$ as a part of the inputs for the BERT response ranker. The action $a_i$ is sampled \footnote{In testing phase, $a_i$ is chosen by the maximum probability.} according to a probability distribution produced by the policy network $\pi (\mathbf{s_i})$ that consists of one feed-forward network as follows:
\begin{equation}
    \pi(\mathbf{s_i}) = P(a_i|\mathbf{s_i}) = softmax(\mathbf{W_{\pi}}^T\mathbf{s}_i + \mathbf{b_{\pi}}),
\end{equation}
where $\mathbf{W_{\pi}}$ and $\mathbf{b_{\pi}}$ are trainable parameters of the policy network.

\subsubsection{Reward Function}
After the selection, we have a subset of PRF terms $\mathcal{P}'$, which is used to update the ranker together with $\mathcal{U}$, $r$ and $y$. 
With the selected PRF terms and the updated BERT response ranker, 
we can conduct inference on the validation data to compute the rewards. 
More specifically, for each batch, the reward function is defined as a delta loss on the validation data:
\begin{equation}
    \mathbf{r}_b = \mathcal{L}_{b-1} -  \mathcal{L}_b,
\end{equation}
where $\mathcal{L}_b$ and $\mathcal{L}_{b - 1}$ are the cross-entropy losses on the validation data for the current batch and the previous batch respectively. The intuition is that if the current loss is smaller than the previous loss, we encourage the selector to follow the current policy by giving it a positive reward. 
Other metrics generated on the validation data could also be incorporated. To enable fast training, we obtain the reward by evaluating a subset of the validation data. This subset is referred to as the \textit{reward set}, which is randomly sampled from the validation data and changed at the end of every episode. 

Furthermore, we compute the future total reward for each batch after an episode since the decisions of the agent not only have a direct impact on the immediate rewards but also have long-term influence. 
The amended reward $\mathbf{r}'$ is formulated as follows.
\begin{equation}
    \mathbf{r}'_b = \sum_{k=0}^{N_e - b} \gamma^k \mathbf{r}_{b+k},
\end{equation}
where $N_e$ is the number of batches in this episode, $\mathbf{r}'_b$ is the future total reward for batch $b$, and $\gamma$ is the reward discount factor.

\subsubsection{Optimization}
We use a policy gradient method, i.e., REINFORCE ~\cite{DBLP:journals/ml/Williams92} to optimize our proposed reinforced PRF Term selector. For a given episode, our goal is to maximize the expected total reward, which can be formulated as follows:
\begin{equation}
    J(\Theta) = E_{\pi_{\Theta}}[\sum_{b=1}^{N_e}\mathbf{r}_b], \label{eq:future_reward}
\end{equation}
where the policy network $\pi_{\Theta}$ is parameterized by $\Theta$. The policy network can be updated by the gradient as follows:
\begin{equation}
    \Theta  \leftarrow \Theta + \alpha \frac{1}{B}\sum_{i=1}^{B} \mathbf{r}_i \nabla_{\Theta}log\pi_{\Theta}(\mathbf{S}_i) \label{eq:policy_gradient}.
\end{equation}
Here, $\alpha$ is the learning rate, $B$ is the batch size. 

\subsection{Training Process}
The selector and the ranker modules are learned jointly as they interact with each other closely during training. 
For each batch, the PRF term selector selects a subset of PRF terms $\mathcal{P}'$ from the input PRF term set $\mathcal{P}$. 
Then the BERT response ranker uses $\mathcal{P}'$ together with the context $\mathcal{U}$ and the response $r$ as inputs to output the predictions. 
To optimize the BERT response ranker, we use a standard gradient descent method to minimize the loss function in Eq. \ref{eq:xent_loss}. 
The reinforced PRF selector intervenes before every iteration of the ranker update by selecting helpful PRF terms to augment the candidate responses. 
Such an intervention process has a direct impact on the gradient computed for the ranker update. 
The BERT response ranker provides a reward in turn to evaluate the utility of the selected PRF terms. 
After each episode, the policy network of the selector is updated with the policy gradient algorithm with the stored (state, action, reward) triples. 
A detailed description of our algorithm is shown in Algorithm \ref{alg:training}

\begin{algorithm}[t]
\caption{Training Procedure}
\begin{algorithmic}[1]
\label{alg:training}
\REQUIRE { 
    \;\\
Training data $\mathcal{D}_{train} = \{ \mathcal{X}_i \}_{i=1}^N =  \{(\mathcal{U}_i, r_i, \mathcal{P}_i, y_i) \}_{i=1}^N$;  \\
Validation data $\mathcal{D}_{val} = \{ \mathcal{X}_i \}_{i=1}^{N'}  = \{(\mathcal{U}_i, r_i, \mathcal{P}_i, y_i) \}_{i=1}^{N'}$; \\
Episode $N_e$, validation sample rate $q$;
}
\STATE Initialize the pre-trained BERT response ranker $g(\cdot)$;
\STATE Initialize the policy network  $\pi_{\Theta}$ as the PRF term selector;

\FOR{episode $l = 1$ to $N_e$} 
    \STATE Obtain the random batch sequence $\mathcal{D}_{train}' = \{ \mathcal{X}_b \}_{b=1}^B$;
    \STATE Obtain the reward set $\mathcal{D}_{reward}$ by random sampling from $\mathcal{D}_{val}$ with rate $q$;
    \FOR{each $\mathcal{X}_b$ in $\{\mathcal{X}_b \} _ {b=1}^B$ }
        \STATE Obtain state $\mathcal{S}_b$ by $z(r_b, \mathcal{P}_b)$;
        \STATE Sample action $\mathcal{A}_b$ according to policy $\pi(\mathcal{S}_b)$;
        \STATE Obtain PRF subset $\mathcal{P}'_b$ according to $\mathcal{A}_b$;
        \STATE Update the ranker $g(\cdot)$ by $\mathcal{X}_b' = \{ \mathcal{U}_b, r_b, \mathcal{P}_b', y_b \}$;
        \STATE Obtain the reward $\mathbf{r}_b$ on $\mathcal{D}_{reward}$;
        \STATE Store ($\mathcal{S}_b$, $\mathcal{A}_b$, $\mathbf{r}_b$) to an episode history $\mathcal{H}$;
    \ENDFOR
    \FOR{each ($\mathcal{S}_b$, $\mathcal{A}_b$, $\mathbf{r}_b$) in $\mathcal{H}$}
    \STATE Obtain the future total reward $\mathbf{r}_b'$ as in Eq. \ref{eq:future_reward};
    \STATE Update the policy network $\pi_{\Theta}$ following Eq. \ref{eq:policy_gradient};
    \ENDFOR
    \STATE Empty $\mathcal{H}$;
\ENDFOR
\end{algorithmic}
\end{algorithm} 
\section{Experiments}
\label{sec:exp}

  \begin{table}[h!]
     \centering
     \caption{The statistics of experimental datasets, where C denotes context and R denotes response. \# Cand. per C denotes the number of candidate responses per context. Note that we didn't filter any stop words or words with low frequency when we computed the average length of contexts or responses.}
     \label{tab:exp_data_stat_train_valid_test}
     \begin{tabular}{l | l l l| l l l}
         \toprule
         Data     & \multicolumn{3}{c|}{MSDialog} & \multicolumn{3}{c}{AliMe-CQA} \\ \hline
         Items  & Train     & Valid   & Test   & Train    & Valid  & Test   \\ \hline
         \# C-R pairs & 173k   & 37k  & 35k  & 32k   & 3.9k  & 4k  \\ 
         \# Cand. per C      & 10        & 10      & 10      & 16.1       & 15     & 15.1     \\ 
         \# + Cand. per C & 1         & 1       & 1       & 6.0      & 5.2   & 4.8    \\ 
         Min \# turns per C & 2 & 2 & 2 & 2 & 2 & 2  \\ 
         Max \# turns per C & 11        & 11      & 11      & 2        & 2      & 2      \\ 
         Avg \# turns per C & 5.0 & 4.9 & 4.4 & 2 & 2 & 2    \\ 
         Avg \# words per C & 451 & 435 & 375 & 17.9 & 18.2 & 18.4  \\ 
         Avg \# words per R & 106 & 107 & 105 & 6.0 & 5.2 & 4.8 \\ \bottomrule
     \end{tabular}
 \end{table}

In this section, we conduct extensive experiments to evaluate the performance of the proposed framework. We also present the online deployment results to show its superiority.

 \subsection{Dataset Description}
 \label{sec:data_desc}

 We evaluate our method and the competing methods on two info-seeking conversation datasets: MSDialog dataset and AliMe-CQA dataset as used in~\cite{DBLP:conf/sigir/YangQQGZCHC18}. The data split and the statistics of the data is shown in Table \ref{tab:exp_data_stat_train_valid_test}.

\subsubsection{MSDialog.}
The MSDialog\footnote{\url{https://ciir.cs.umass.edu/downloads/msdialog/}} dataset is a labeled dialog dataset of question answering (QA) interactions between information seekers and answer providers from an online forum on Microsoft products ~\cite{DBLP:conf/sigir/QuYCTZQ18}. Previous works \cite{DBLP:conf/sigir/YangQQGZCHC18} have a pre-processed version that is suitable for experimenting with conversation response ranking models. The ground truth responses returned by the real agents are the positive response candidates, and negative sampling has been adopted to create nine negative response candidates for each context query. 
We only removed some common prefixes such as ``\texttt{<<AGENT>>:}'' and use WordPiece ~\cite{DBLP:journals/corr/WuSCLNMKCGMKSJL16} to tokenize the context and the response candidates for further modeling.

\subsubsection{AliMe-CQA Data.}
We collected the multi-turn question answering chat logs between customers and a chatbot from the AliMe conversation system ~\footnote{https://www.alixiaomi.com/}. This chatbot is built based on a question-to-question matching system~\cite{alime-demo}, where for each query, it finds the most similar candidate question in a QA database and returns its answer as the reply. To form an information-seeking conversation QA dataset, we firstly select more than 3k multi-turns context to form queries and apply this conversation system to retrieve the top-15 most similar candidate questions as the ``response'' in our setting. A group of business analysts is asked to annotate the candidate ``response''. If the ``response'' is similar to the input query (context), the label will be positive, otherwise negative. In the process of annotation, if the confidence score of answering a given query (context) is low, the system will prompt three top related questions(response candidates) for users to choose from. We collected such user click logs as our external data, where we treat the clicked question as positive and the others as negative. We have recalled about 50k context-response pairs from this annotation process and remove all of the contexts that have zero positive candidate responses. The language of the context and response is Chinese and we use character-level tokenization for further modeling.

\subsection{PRF Term Candidates Retrieving}
The goal of pseudo-relevance feedback (PRF) is to extract terms from the top-ranked documents during the retrieval process to help discriminate relevant documents from the irrelevant ones ~\cite{Cao:2008:SGE:1390334.1390377}. 

The expansion terms are extracted either according to the term distributions (e.g., extract the most frequent terms) or extracted from the most specific terms (e.g., extract terms with the maximum IDF weights) in feedback documents. Given the retrieved top $K_1$ QA posts $\mathcal{P}$ from the previous step, we compute a language model $\theta=P \left(w|\mathcal{P} \right)$ using $\mathcal{P}$. Then we extract the most frequent $K_2$ terms from $\theta$ as the PRF terms for the response candidate.
We first use the response candidate as the query to retrieve top $K_1$ QA posts from the external corpus with BM25 as the source of the external knowledge for each response candidate. While retrieving relevant documents, we perform several preprocessing steps including tokenization, punctuation removal, and stop word removal for the query. We set $K_1=10$ for MSDialog dataset, and $K_1=15$ for AliMe dataset. The external retrieval source corpus for MSDialog and AliMe are Stack Overflow data and unlabeled AliMe QA data respectively.

Once the response-relevant QA posts are retrieved for each response, we count the term frequencies for all terms in these posts and select the top $K_2$ frequent terms as the PRF terms for each response candidate. $K_2$ is set as $10$ for both the MSDialog dataset and the AliMe dataset.

\begin{table*}[th]
    \centering
    \caption{Comparison of different models over MSDialog and eCommerce data sets.  Numbers in bold font mean the result is the best compared with other models. } 
    \label{tab:baseline_results}
    \begin{tabular}{l|c c c c |c c c c c}
        \toprule
        Data  & \multicolumn{4}{c|}{MSDialog}           & \multicolumn{5}{c}{AliMe-CQA}          \\ \midrule
        Methods    & Recall@1 & Recall@2 & Recall@5 & MAP    & Precision@1 & Recall@1 & Recall@2 & Recall@5 & MAP    \\ 
        \texttt{BM25} \cite{Robertson:1994:SEA:188490.188561}    & 0.2626   & 0.3933   & 0.6329   & 0.4387 & 0.5811 & 0.2012   & 0.3201   & 0.5378   & 0.6310 \\  
        \texttt{ARC-II} \cite{DBLP:conf/nips/HuLLC14} & 0.3189   & 0.5413   & 0.8662   & 0.5398 & 0.6075 & 0.1717   & 0.3027   & 0.6190   & 0.6841 \\  
        \texttt{MV-LSTM} \cite{DBLP:conf/aaai/WanLGXPC16}  & 0.2768   & 0.5000   & 0.8516   & 0.5059 & 0.5925 & 0.1657   & 0.3194   & 0.6015   & 0.6813 \\ 
        \texttt{DRMM} \cite{Guo:2016:DRM:2983323.2983769} & 0.3507   & 0.5854   & 0.9003   & 0.5704 & 0.6868 & 0.2194   & 0.3563   & 0.6036   & 0.7048 \\  
        \texttt{Duet} \cite{Mitra:2017:LMU:3038912.3052579} & 0.2934   & 0.5046   & 0.8481   & 0.5158 & 0.6679 &  0.1920   & 0.3408   & 0.6302   & 0.7162 \\ 
        \texttt{DAM} \cite{DBLP:conf/acl/WuLCZDYZL18}     & 0.7012   & 0.8527   & 0.9715   & 0.8150 & 0.7558 &  0.2472  & 0.3969   &  0.6919  & 0.7773 \\ \hline
        \texttt{BERT-Ranker} & 0.7667 & 0.8926 & \textbf{0.9852} & 0.8580 & 0.8476 & 0.2968 & 0.4622 & 0.7263 & 0.8513    \\ 
        \texttt{PRF-RL} & \textbf{0.7872} & \textbf{0.9032} & 0.9792 & \textbf{0.8700} & \textbf{0.8717} & \textbf{0.3181} & \textbf{0.4868} & \textbf{0.7576} & \textbf{0.8675}  \\ 
        \bottomrule
    \end{tabular}
\end{table*}

\subsection{Experimental Setup}
\subsubsection{Baselines}
We explore different baselines lying on four categories, including traditional retrieval models, neural ranking models, a strong multi-turn conversation response ranking method, pre-trained language model based models as follows:

\textbf{BM25} ~\cite{Robertson:1994:SEA:188490.188561} is a traditional retrieval model, which uses the dialog context as the query to retrieve response candidates for response ranking.

\textbf{ARC-II} ~\cite{DBLP:conf/nips/HuLLC14}, \textbf{MV-LSTM} ~\cite{DBLP:conf/aaai/WanLGXPC16}, \textbf{DRMM} ~\cite{Guo:2016:DRM:2983323.2983769}, \textbf{DUET} ~\cite{Mitra:2017:LMU:3038912.3052579} are neural ranking models proposed in recent years for ad-hoc retrieval and question answering. MV-LSTM is a representation focused model and ARC-II, DRMM are interaction focused models. Duet is a hybrid method of both representation focused and interaction-focused models.

\textbf{DAM} ~\cite{DBLP:conf/acl/WuLCZDYZL18} is a strong baseline model for response ranking in multi-turn conversations. DAM also represents and matches a response with its multi-turn context using dependency information learned by Transformers. 

\textbf{BERT-Ranker} is a general classification framework proposed in BERT ~\cite{DBLP:conf/naacl/DevlinCLT19} paper. It uses \texttt{[SEP]} and segment embedding to separate the query and answer and incorporates a pre-trained language model for contextual representation. The predictions are based on the contextual vector of \texttt{[CLS]} token.

\textbf{PRF-RL} is our proposed model consisting of a reinforced PRF term selector and a BERT response ranker.

\subsubsection{Evaluation Methodology}
For evaluation metrics of both MSDialog and AliMe-CQA, we adopted mean average precision (MAP) and Recall@k which is the recall at top $k$ ranked responses from n available candidates for a given conversation context.  Following previous related works ~\cite{DBLP:conf/acl/WuLCZDYZL18}, here we reported Recall@1, Recall@2, and Recall@5 on both two datasets. For AliMe-CQA, we reported an extra metric Precision@1 for further exploration, since there are multiple positive candidates of a given query. One should also notice that for the MSDialog dataset, the value of precision@1 is equal to the recall@1 since there is only one positive candidate for each query in this dataset.

\subsubsection{Experimental Settings}
The four neural ranking models are experimented using the MatchZoo ~\footnote{
https://github.com/NTMC-Community/MatchZoo} toolkit. 
We use the code ~\footnote{https://github.com/baidu/Dialogue/tree/master/DAM}  released by  \cite{DBLP:conf/acl/WuLCZDYZL18} to tune the DAM model on our datasets. We use the Hugging Face transformer version ~\footnote{https://github.com/huggingface/transformers} of the BERT model to implement BASE-BERT classifier.  We choose the $\textbf{BERT}_{\textbf{BASE}}$ (L=12, H=768, A=12)
 as our pre-trained BERT encoder for both BERT-Ranker and PRF-RL. The models are implemented in PyTorch and run on 1 Tsela P100 GPU. We present the detailed hyper-parameters used for experiments as follows.

For the MSDialog dataset, the context length is truncated by 384 and the response length is truncated by 96. The batch size is set to 12. We use Adam optimizer with linear decay for both two models. The learning rate for BERT-Ranker is set to 3e-5 following the previous works.  For PRF-RL, we firstly pre-trained the BERT response ranker without PRF term selection of learning rate 3e-5 for 1000 steps, and then jointly trained the reinforced PRF term selector and the BERT response ranker with learning rate 1e-4, 1.5e-5 respectively. For reinforcement learning, we use max-pooling in the state representation, the number of the episode is set to 100, the reward discount reward factor is set to 0.3, and the reward set is randomly sampled from 0.5\% of the validation dataset considering the trade-off of quality and efficiency of training.

For the AliMe dataset, the context length is truncated by 100 and the response length is truncated by 50. The batch size is set to 32. We again use Adam optimizer with linear decay for both two models. The learning rate for BERT-Ranker is set to 3e-5. For PRF-RL, we firstly pre-trained the BERT response ranker with a learning rate 3e-5 for 200 steps, and then jointly trained the reinforced PRF term selector and the BERT response ranker with a learning rate 1e-4, 5e-6 respectively. We again use max-pooling in the state representation, the number of the episode is set to 10, the reward discount reward factor is set to 0.3, and the reward set is randomly sampled from 1\% of the validation dataset.

\subsection{Comparison with Baselines}
We present evaluation results over different methods on MSDialog and AliMe-CQA in Table \ref{tab:baseline_results}.

\subsubsection{Performance Comparison on MSDialog}
From the results on the MSDialog dataset, we have the following findings.
First, the transformer-based models (DAM, BERT-Ranker, PRF-RL) show significant improvements compared with traditional retrieval models and other neural ranking models, which further proves the powerful representation capabilities of \haojie{the} Transformer.  Second, compared with the DAM model, when the pre-trained language model BERT is incorporated, the performance also has a good improvement. Last but not least, our model performs the best over all the other baselines on Recall@1, Recall@2, and MAP, which we can see that incorporating external knowledge via pseudo-relevance feedback could improve the performance of the BERT-based response ranking models by large margins. Specifically, compared with BERT-Ranker, our proposed PRF-RL model has a comparable result in terms of Recall@5, but an improvement of 2.05\% for Recall@1, 1.06\% for Recall@2, 1.20\% for MAP. This shows the benefits of considering PRF selection.


\subsubsection{Performance Comparison on AliMe-CQA}
After comparing our PRF-RL model with other baselines on the AliMe-CQA dataset in Table~\ref{tab:baseline_results}. We find those similar findings as using MSDialog. First, our model achieves the best performance against all the baselines in terms of all evaluation metrics. Specifically, for precision-based metrics, our model achieves 2.41\% and 1.62\% improvements compared with the strongest baseline BERT-Ranker in terms of Precision@1 and MAP; And for recall-based metrics, our model achieves improvements of 2.13\% for Recall@1, 2.46\% for Recall@2 and 3.13\% for Recall@5. Second, Compared with the MSDialog dataset, the absolute values of Recall@k are lower. This phenomenon comes from multiple positive candidates given one query, in such case, a lower recall comparing with MSDialog dataset does not necessarily mean the method has lower performance. 
In practice, for info-seeking conversation systems, Precision@1 is the most important metric as only the top-1 response will be returned to the customer. In this metric, all the methods on AliMe-CQA tend to have better performance than MSDialog. 

In all, our proposed method has a clear advantage over all the competing methods in both datasets, which demonstrates the usefulness of our method for info-seeking conversations.

\subsection{Comparison with Other PRF Methods}

To further explore how well our reinforced PRF term selector contributes to the overall model performance, we build several baseline methods that use different ways to incorporate the information of pseudo-relevance feedbacks. Here we have the following baselines:

\textbf{RULE-PRF} is a simple method to feed the PRF terms filtered by term frequencies, which is introduced in ~\cite{DBLP:conf/sigir/YangQQGZCHC18}. 

\textbf{PRF-ML-Tanh} is a soft selection method which outputs a score $q_{p_i} = \tanh{(\mathbf{s}_i)}$. Before the embedding of PRF terms $p_i$ is feed into BERT, the embedding will first be scaled by this score $q_{p_i}$. It is more like a gate function to ensure that the gradients can be back-propagated into the selector.

\textbf{PRF-ML-Sig} is a similar soft selection method which replaces the $tanh$ function with the $sigmoid$ function.

\textbf{PRF-ML-Gumb} is a method that can not only produce a hard selection decision but also can be optimized by gradient descent based algorithms. It uses a categorical re-parameterization trick with the Gumbel softmax function ~\cite{DBLP:conf/iclr/JangGP17} that enables the model to sample discrete random variables in a way that is differentiable. 

\begin{table}[t]
    \centering
    \caption{Comparison of different models over MSDialog and AliMe-CQA datasets.  Numbers in bold font mean the result is the best compared with other models. Here ``P'' means precision and ``R'' means recall. The Precision@1 results on the MSDialog dataset are omitted since it is equal to Recall@1.} 
    \label{tab:comparision_prf_methods}
    \begin{tabular}{l|c c c c c }
        \toprule
        \multicolumn{6}{c}{MSDialog}    \\ \midrule
        Methods  & P@1  & R@1 & R@2 & R@5 & MAP \\ 
        \texttt{BERT-Ranker} & / & 0.7667 & 0.8926 & \textbf{0.9852} & 0.8580 \\ 
        \texttt{RULE-PRF} & / & 0.7713  & 0.8906 & 0.9826 & 0.8601\\ 
        \texttt{PRF-ML-Tanh} & / & 0.7770 & 0.8886 & 0.9815 & 0.8626\\
        \texttt{PRF-ML-Sigmoid} & / & 0.7736 & 0.8906 & 0.9823 & 0.8607\\  
        \texttt{PRF-ML-Gumbel} & / & 0.7719 & 0.8946 & 0.9823 & 0.8614\\ 
        \texttt{PRF-RL} & / & \textbf{0.7872} & \textbf{0.9032} & 0.9792 & \textbf{0.8700}\\ 
        \bottomrule
    \end{tabular}

    \begin{tabular}{l|c c c c c}
        \multicolumn{6}{c}{AliMe-CQA}          \\ \midrule
        Methods    & P@1 & R@1 & R@2 & R@5 & MAP    \\ 
        \texttt{BERT-Ranker} & 0.8476 & 0.2968 & 0.4622 & 0.7263 & 0.8513    \\ 
        \texttt{RULE-PRF}  & 0.8460 & 0.3002 & 0.4785 & 0.7422 & 0.8523  \\ 
        \texttt{PRF-ML-Tanh} & 0.8604 & 0.3060 & 0.5027 & 0.7631 & 0.8654   \\
        \texttt{PRF-ML-Sigmoid} &  0.8340 & 0.3041 & 0.4835 & 0.7362 & 0.8549  \\  
        \texttt{PRF-ML-Gumbel} & 0.8566 & 0.3084 & 0.4781 & 0.7429 & 0.8554   \\ 
        \texttt{PRF-RL} &  \textbf{0.8717} & \textbf{0.3181} & \textbf{0.4868} & \textbf{0.7576} & \textbf{0.8675}  \\ 
        \bottomrule
    \end{tabular}
\end{table}

The experimental results are shown in Table~\ref{tab:comparision_prf_methods}. \haojie{By exploring the results,} we have the following findings: 

$\bullet $ Overall, the incorporation of pseudo-relevance feedback can improve the performance of the BERT-based response ranking models. The RULE-PRF method without selection achieves improvements of 0.46\% for Recall@1 on the MSDialog dataset and 0.34\% for Recall@1 on the AliMe-CQA dataset. However, in terms of some metrics such as Recall@2 and Recall@5 on MSDialog dataset, and Precision@1 on the AliMe-CQA dataset, adding PRF terms can hurt the ranking model's performance. This means the necessity to implement better ways for PRF term selection, instead of the simple approach. 

$\bullet $ For a soft version of machine learning based PRF term selection models, the tanh gating is better than sigmoid gating. Hard version (PRF-ML-Gumb) of machine learning based PRF term selection performance better in terms of Recall@2, Recall@5 on MSDialog Dataset and Recall@1 on AliMe-CQA dataset, which can be concluded that the hard selection has potential to achieve better performance but training hard selection by machine learning is not straightforward and intuitive. 

$\bullet $ After incorporating the hard selection by our reinforced PRF term selector, our proposed PRF-RL model achieves the best performance against outperforms all the PRF selection methods. This observation again proves the effectiveness of our framework in PRF term selection. 

\subsection{Hyper-parameter Sensitivity Analysis}

\begin{figure}[t]
\centering
\includegraphics[width=0.95\linewidth]{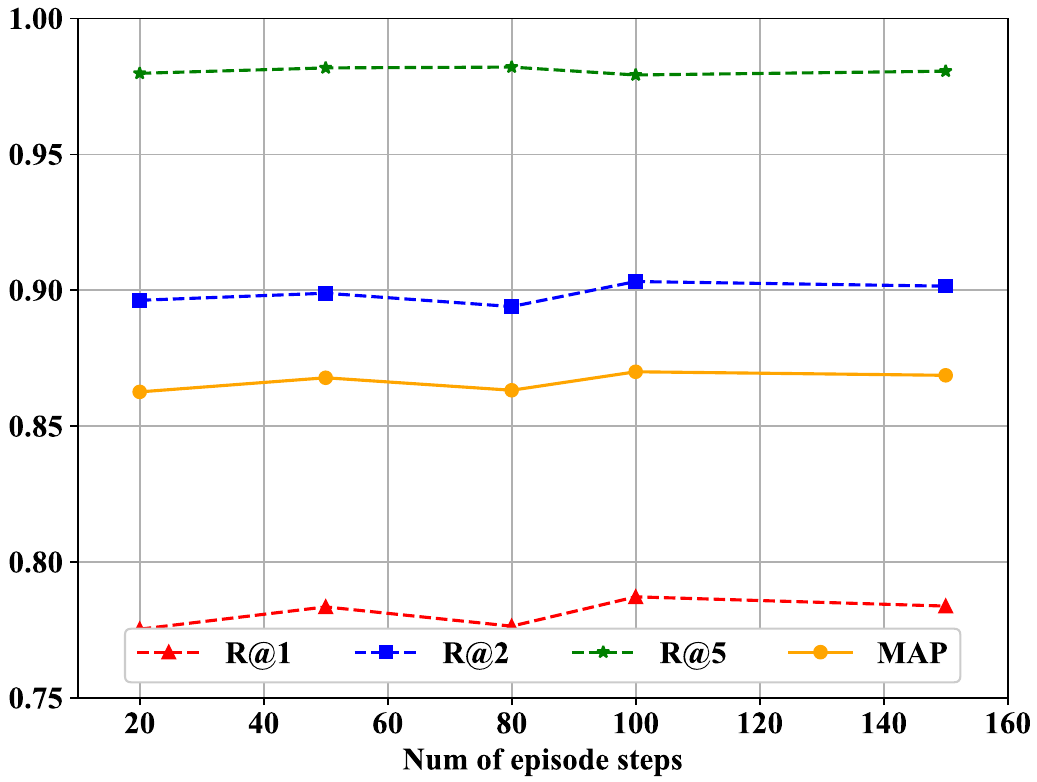}
\includegraphics[width=0.95\linewidth]{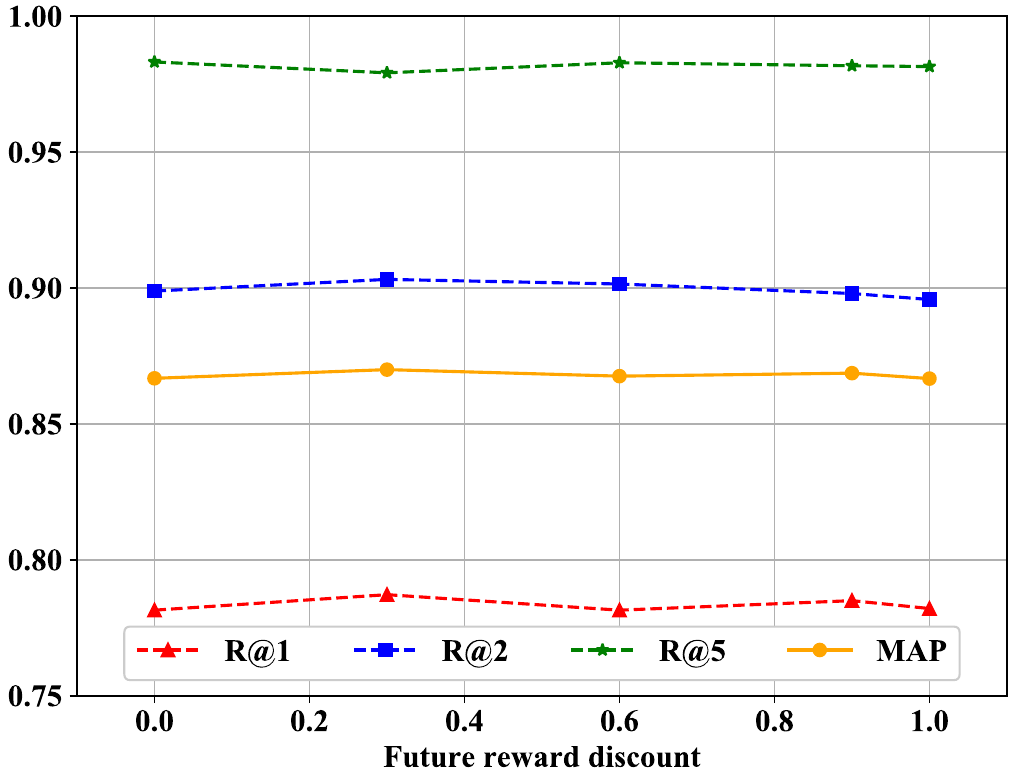}
\caption{Performance of PRF-RL with different choices of numbers of episodes and discount factors over the test partition of MSDialog. The reward discount factor is set to 0.3 when we compare different choices of the number of episodes (Left). The number of episodes is set to 100 while we compare different choices of the discount factors (Right).}
\label{fig:ablation_study}
\end{figure}

In this section, we first examine the performance of PRF-RL with different choices of the number of episode steps on the MSDialog dataset. As in Figure \ref{fig:ablation_study} (Left), we have these observations. First, we find that our method is generally insensitive to this parameter, although by setting the number of episode steps as 100, our method has a slight improvement.
Second, we observe that the performance in terms of Recall@5 is inconsistent with other metrics such as Recall@1 and Recall@2. A good performance in Recall@5 may not come with good performances in Recall@1 and Recall@2. This means the model optimizes the metrics differently, and may not achieve the best performance on all these metrics. In practice, Recall@1 is more important than Recall@2 or Recall@5. Thus we can set the episode step as 100 as we can achieve the best Recall@1, Recall@2, and MAP. 
We then proceed to examine our model performance w.r.t. the future reward discount factors on the MSDialog dataset in Figure~\ref{fig:ablation_study} (Right). In general, the model performance is not very sensitive to this parameter as well, although our method has slightly better performance in terms of Recall@1 and Recall@2 by setting the reward discount factor as 0.3. 
From Figure~\ref{fig:ablation_study}, we find our method is pretty robust as it is not very sensitive to these parameter settings. But still, by conducting combining the findings in both Figures, we can find a relatively better combination of hyper-parameters for our method.

\begin{figure*}[t!]
\centering
\includegraphics[width=0.98\linewidth]{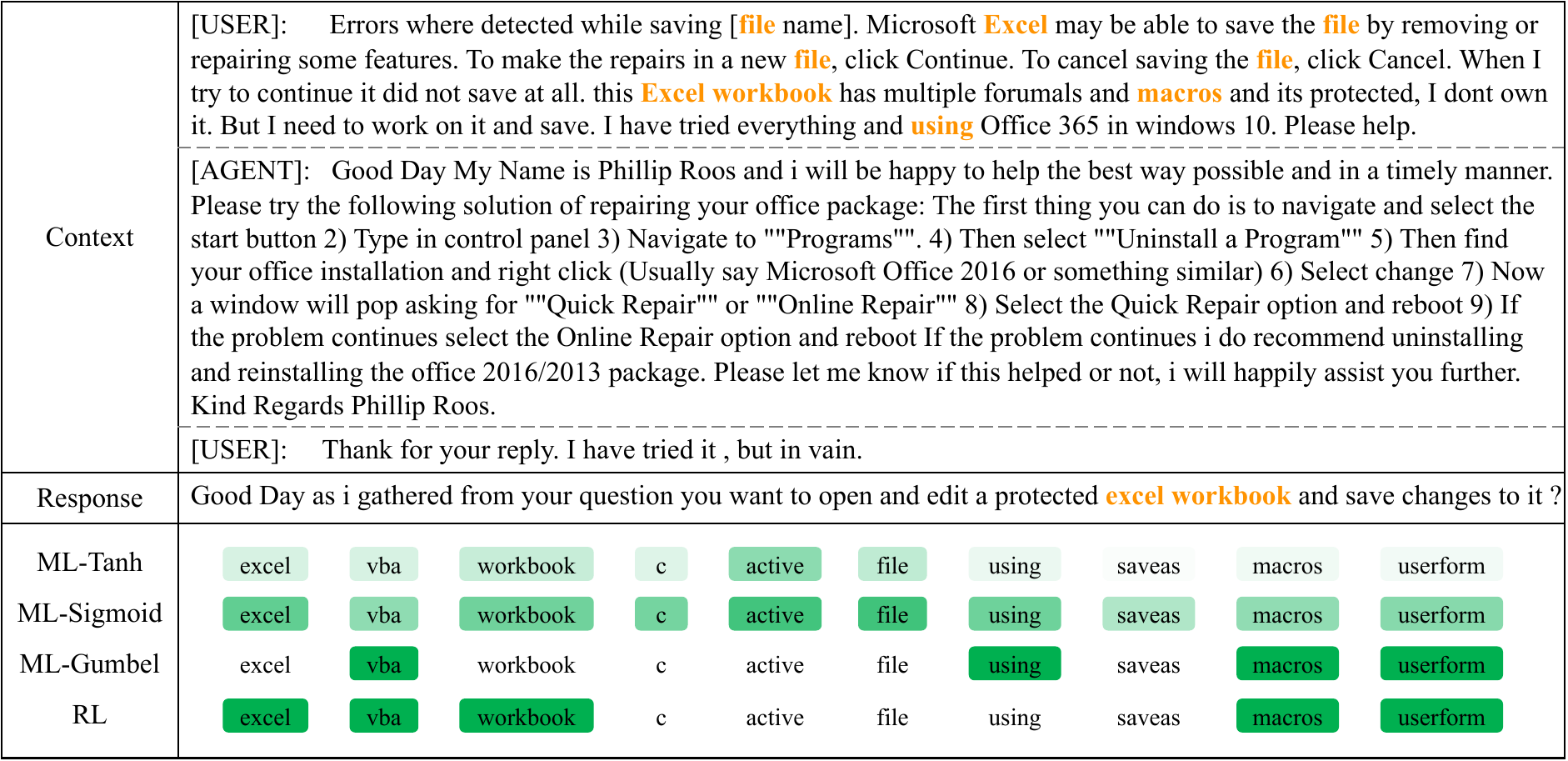}
\caption{Examples of PRF selection using different models.  Here, the response is a positive candidate and we use different levels of transparency of {\color{mgreen}green color} to demonstrate the levels of PRF term selection for expanding this response. The {\color{morange}\textbf{bold orange}} terms are the PRF terms shown in context/response, sorted by their term frequencies.}
\label{fig:case_study}
\end{figure*}

\subsection{Case Study}
One of the most interesting features of our model is that it exhibits a certain level of interpretability as the selection process is explicit. 
We then present a case study to examine whether the generated PRF terms for expanding the response candidates are meaningful.

Figure~\ref{fig:case_study} shows one of the cases generated from the MSDialog dataset. In this case, the user wants to edit and save a ``excel notebook'' which seems to be protected by multiple formulas and macros. The agent proposes a general solution to the user but not working. So the response might be other solutions or ask the user about more detailed descriptions of the problem he has met. Since the ``excel notebook'' can be converted to ``protected'' one through some ``macros'', the contextual correlation between ``protected excel notebook'' with ``macros'' and ``vba'' is strong. From the results, we can find the soft selections of ML-Tanh and ML-Sigmoid can confuse the ranking model since they both give more weights on irrelevant terms such as ``c'' and ``active''. ML-Gumbel has the same problem, it selects a general term ``using'' and dropped the ``excel'' and ``workbook'' which are relevant to the response. Our PRF-RL model achieve the best result since it can (1) select the exact match terms such as ``excel'' and ``workbook'' in the response, (2) avoid selecting irrelevant or general terms such as ``c'', ``active'' and ``using'', (3) select contextually correlated terms such as ``macros'', which appears in the context but not in the response, which means it can be used to improve the recall of the response candidate. In all, encouragingly we find the selection process made by the proposed PRF-RL model is insightful and intuitive.

\subsection{Online A/B Test}

Finally, we deployed the proposed PRF-RL model on an online chatbot engine called AliMe~\footnote{\url{https://www.alixiaomi.com/}} in the e-commerce company Alibaba, and conduct an A/B test on our proposed model and the existing online ranking system without considering external PRF expansion. In the AliMe chatbot engine, for each user query or request, it employs a two-stage model to find a suitable candidate response. It first calls back at most 15 candidate responses from tens of thousands of candidates and then uses a neural ranker to rerank the candidate responses. The engine uses both of the two systems, one is our method and the other is the online method, to rerank the candidates. Note that the online method is a degenerated version of our method without considering PRF expansion.
Overall we have randomly selected 13,549 conversational QA-pairs. After filtering out all the context queries that have zero positive responses in the call-back set, we have collected 325 context queries for each system. We then ask a customer agent to annotate the results of both methods. We obtain the number of the hit of the top-1 ranked candidates and compared the precision@1 score. 

As a result, our proposed PRF-RL model has Precision@1 of 67.69\%, which has a significant relative improvement (12.24\%) compared with the existing online ranking system (60.3\%). Such improvement is considered to be a big improvement for the chatbot engine. This further shows the advantage of our proposed method and the usefulness of the external PRF expansion. For efficiency, the candidate responses are expanded in a pre-processing step which is done in an offline manner, thus it ensures zero efficiency drop.


\section{Conclusion} \label{sec:conlusion}
In this work, we propose a principled approach to automatically select useful pseudo-relevance feedback terms to help information-seeking conversations. 
The proposed method considers a reinforced selector to interact with a BERT response ranker to generate high-quality pseudo-relevance feedback terms, and the performance of the ranker can help to guide the behaviors of the selector. 
Extensive experiments on both public and industrial datasets show our model outperforms the competing models. 
We have also deployed our proposed model in AliMe chatbot and observe a large improvement over the existing online ranking system.

\bibliographystyle{ACM-Reference-Format}
\bibliography{sample-base}


\begin{thebibliography}{57}


\ifx \showCODEN    \undefined \def \showCODEN     #1{\unskip}     \fi
\ifx \showDOI      \undefined \def \showDOI       #1{#1}\fi
\ifx \showISBNx    \undefined \def \showISBNx     #1{\unskip}     \fi
\ifx \showISBNxiii \undefined \def \showISBNxiii  #1{\unskip}     \fi
\ifx \showISSN     \undefined \def \showISSN      #1{\unskip}     \fi
\ifx \showLCCN     \undefined \def \showLCCN      #1{\unskip}     \fi
\ifx \shownote     \undefined \def \shownote      #1{#1}          \fi
\ifx \showarticletitle \undefined \def \showarticletitle #1{#1}   \fi
\ifx \showURL      \undefined \def \showURL       {\relax}        \fi
\providecommand\bibfield[2]{#2}
\providecommand\bibinfo[2]{#2}
\providecommand\natexlab[1]{#1}
\providecommand\showeprint[2][]{arXiv:#2}

\bibitem[\protect\citeauthoryear{Arulkumaran, Deisenroth, Brundage, and
  Bharath}{Arulkumaran et~al\mbox{.}}{2017}]%
        {DBLP:journals/spm/ArulkumaranDBB17}
\bibfield{author}{\bibinfo{person}{Kai Arulkumaran},
  \bibinfo{person}{Marc~Peter Deisenroth}, \bibinfo{person}{Miles Brundage},
  {and} \bibinfo{person}{Anil~Anthony Bharath}.}
  \bibinfo{year}{2017}\natexlab{}.
\newblock \showarticletitle{Deep Reinforcement Learning: {A} Brief Survey}.
\newblock \bibinfo{journal}{\emph{{IEEE} Signal Process. Mag.}}
  \bibinfo{volume}{34}, \bibinfo{number}{6} (\bibinfo{year}{2017}),
  \bibinfo{pages}{26--38}.
\newblock


\bibitem[\protect\citeauthoryear{Cao, Nie, Gao, and Stephen}{Cao
  et~al\mbox{.}}{2008}]%
        {Cao:2008:SGE:1390334.1390377}
\bibfield{author}{\bibinfo{person}{G. Cao}, \bibinfo{person}{J. Nie},
  \bibinfo{person}{J. Gao}, {and} \bibinfo{person}{R. Stephen}.}
  \bibinfo{year}{2008}\natexlab{}.
\newblock \showarticletitle{Selecting Good Expansion Terms for Pseudo-relevance
  Feedback}. In \bibinfo{booktitle}{\emph{SIGIR '08}}.
\newblock


\bibitem[\protect\citeauthoryear{Devlin, Chang, Lee, and Toutanova}{Devlin
  et~al\mbox{.}}{[n.d.]}]%
        {DBLP:conf/naacl/DevlinCLT19}
\bibfield{author}{\bibinfo{person}{Jacob Devlin}, \bibinfo{person}{Ming{-}Wei
  Chang}, \bibinfo{person}{Kenton Lee}, {and} \bibinfo{person}{Kristina
  Toutanova}.} \bibinfo{year}{[n.d.]}\natexlab{}.
\newblock \showarticletitle{{BERT:} Pre-training of Deep Bidirectional
  Transformers for Language Understanding}. In
  \bibinfo{booktitle}{\emph{NAACL-HLT}}. \bibinfo{pages}{4171--4186}.
\newblock


\bibitem[\protect\citeauthoryear{Fan, Tian, Qin, Bian, and Liu}{Fan
  et~al\mbox{.}}{2017}]%
        {DBLP:journals/corr/FanTQBL17}
\bibfield{author}{\bibinfo{person}{Yang Fan}, \bibinfo{person}{Fei Tian},
  \bibinfo{person}{Tao Qin}, \bibinfo{person}{Jiang Bian}, {and}
  \bibinfo{person}{Tie{-}Yan Liu}.} \bibinfo{year}{2017}\natexlab{}.
\newblock \showarticletitle{Learning What Data to Learn}.
\newblock \bibinfo{journal}{\emph{CoRR}} (\bibinfo{year}{2017}).
\newblock


\bibitem[\protect\citeauthoryear{Fang, Li, and Cohn}{Fang
  et~al\mbox{.}}{2017}]%
        {DBLP:journals/corr/abs-1708-02383}
\bibfield{author}{\bibinfo{person}{Meng Fang}, \bibinfo{person}{Yuan Li}, {and}
  \bibinfo{person}{Trevor Cohn}.} \bibinfo{year}{2017}\natexlab{}.
\newblock \showarticletitle{Learning how to Active Learn: {A} Deep
  Reinforcement Learning Approach}.
\newblock \bibinfo{journal}{\emph{CoRR}} (\bibinfo{year}{2017}).
\newblock


\bibitem[\protect\citeauthoryear{Feng, Huang, Zhao, Yang, and Zhu}{Feng
  et~al\mbox{.}}{2018}]%
        {DBLP:conf/aaai/FengHZYZ18}
\bibfield{author}{\bibinfo{person}{Jun Feng}, \bibinfo{person}{Minlie Huang},
  \bibinfo{person}{Li Zhao}, \bibinfo{person}{Yang Yang}, {and}
  \bibinfo{person}{Xiaoyan Zhu}.} \bibinfo{year}{2018}\natexlab{}.
\newblock \showarticletitle{Reinforcement Learning for Relation Classification
  From Noisy Data}. In \bibinfo{booktitle}{\emph{AAAI}}.
  \bibinfo{publisher}{{AAAI} Press}, \bibinfo{pages}{5779--5786}.
\newblock


\bibitem[\protect\citeauthoryear{Gao, Galley, and Li}{Gao
  et~al\mbox{.}}{2018}]%
        {DBLP:journals/corr/abs-1809-08267}
\bibfield{author}{\bibinfo{person}{J. Gao}, \bibinfo{person}{M. Galley}, {and}
  \bibinfo{person}{L. Li}.} \bibinfo{year}{2018}\natexlab{}.
\newblock \showarticletitle{Neural Approaches to Conversational {AI}}.
\newblock \bibinfo{journal}{\emph{CoRR}}  \bibinfo{volume}{abs/1809.08267}
  (\bibinfo{year}{2018}).
\newblock


\bibitem[\protect\citeauthoryear{Guo, Fan, Ai, and Croft}{Guo
  et~al\mbox{.}}{2016}]%
        {Guo:2016:DRM:2983323.2983769}
\bibfield{author}{\bibinfo{person}{J. Guo}, \bibinfo{person}{Y. Fan},
  \bibinfo{person}{Q. Ai}, {and} \bibinfo{person}{W.~B. Croft}.}
  \bibinfo{year}{2016}\natexlab{}.
\newblock \showarticletitle{A Deep Relevance Matching Model for Ad-hoc
  Retrieval}. In \bibinfo{booktitle}{\emph{CIKM '16}}.
\newblock


\bibitem[\protect\citeauthoryear{Henderson, Vulic, Gerz, Casanueva,
  Budzianowski, Coope, Spithourakis, Wen, Mrksic, and Su}{Henderson
  et~al\mbox{.}}{2019}]%
        {DBLP:conf/acl/HendersonVGCBCS19}
\bibfield{author}{\bibinfo{person}{Matthew Henderson}, \bibinfo{person}{Ivan
  Vulic}, \bibinfo{person}{Daniela Gerz}, \bibinfo{person}{I{\~{n}}igo
  Casanueva}, \bibinfo{person}{Pawel Budzianowski}, \bibinfo{person}{Sam
  Coope}, \bibinfo{person}{Georgios Spithourakis},
  \bibinfo{person}{Tsung{-}Hsien Wen}, \bibinfo{person}{Nikola Mrksic}, {and}
  \bibinfo{person}{Pei{-}Hao Su}.} \bibinfo{year}{2019}\natexlab{}.
\newblock \showarticletitle{Training Neural Response Selection for
  Task-Oriented Dialogue Systems}. In \bibinfo{booktitle}{\emph{ACL}}.
  \bibinfo{pages}{5392--5404}.
\newblock


\bibitem[\protect\citeauthoryear{Hu, Lu, Li, and Chen}{Hu
  et~al\mbox{.}}{2014}]%
        {DBLP:conf/nips/HuLLC14}
\bibfield{author}{\bibinfo{person}{B. Hu}, \bibinfo{person}{Z. Lu},
  \bibinfo{person}{H. Li}, {and} \bibinfo{person}{Q. Chen}.}
  \bibinfo{year}{2014}\natexlab{}.
\newblock \showarticletitle{Convolutional Neural Network Architectures for
  Matching Natural Language Sentences}. In \bibinfo{booktitle}{\emph{NIPS
  '14}}.
\newblock


\bibitem[\protect\citeauthoryear{Jang, Gu, and Poole}{Jang
  et~al\mbox{.}}{2017}]%
        {DBLP:conf/iclr/JangGP17}
\bibfield{author}{\bibinfo{person}{Eric Jang}, \bibinfo{person}{Shixiang Gu},
  {and} \bibinfo{person}{Ben Poole}.} \bibinfo{year}{2017}\natexlab{}.
\newblock \showarticletitle{Categorical Reparameterization with
  Gumbel-Softmax}. In \bibinfo{booktitle}{\emph{ICLR}}.
\newblock


\bibitem[\protect\citeauthoryear{Lan, Chen, Goodman, Gimpel, Sharma, and
  Soricut}{Lan et~al\mbox{.}}{2019}]%
        {DBLP:journals/corr/abs-1909-11942}
\bibfield{author}{\bibinfo{person}{Zhenzhong Lan}, \bibinfo{person}{Mingda
  Chen}, \bibinfo{person}{Sebastian Goodman}, \bibinfo{person}{Kevin Gimpel},
  \bibinfo{person}{Piyush Sharma}, {and} \bibinfo{person}{Radu Soricut}.}
  \bibinfo{year}{2019}\natexlab{}.
\newblock \showarticletitle{{ALBERT:} {A} Lite {BERT} for Self-supervised
  Learning of Language Representations}.
\newblock \bibinfo{journal}{\emph{CoRR}} (\bibinfo{year}{2019}).
\newblock


\bibitem[\protect\citeauthoryear{Lavrenko and Croft}{Lavrenko and
  Croft}{2001}]%
        {Lavrenko:2001:RBL:383952.383972}
\bibfield{author}{\bibinfo{person}{V. Lavrenko} {and} \bibinfo{person}{W.~B.
  Croft}.} \bibinfo{year}{2001}\natexlab{}.
\newblock \showarticletitle{Relevance Based Language Models}. In
  \bibinfo{booktitle}{\emph{SIGIR '01}}.
\newblock


\bibitem[\protect\citeauthoryear{Li, Sun, He, Wang, Hui, Yates, Sun, and Xu}{Li
  et~al\mbox{.}}{2018}]%
        {DBLP:conf/emnlp/LiSHWHYSX18}
\bibfield{author}{\bibinfo{person}{Canjia Li}, \bibinfo{person}{Yingfei Sun},
  \bibinfo{person}{Ben He}, \bibinfo{person}{Le Wang}, \bibinfo{person}{Kai
  Hui}, \bibinfo{person}{Andrew Yates}, \bibinfo{person}{Le Sun}, {and}
  \bibinfo{person}{Jungang Xu}.} \bibinfo{year}{2018}\natexlab{}.
\newblock \showarticletitle{{NPRF:} {A} Neural Pseudo Relevance Feedback
  Framework for Ad-hoc Information Retrieval}. In
  \bibinfo{booktitle}{\emph{EMNLP}}.
\newblock


\bibitem[\protect\citeauthoryear{Li, Qiu, Chen, Wang, Gao, Huang, Ren, Zhao,
  Zhao, Wang, and Jin}{Li et~al\mbox{.}}{2017}]%
        {alime-demo}
\bibfield{author}{\bibinfo{person}{F. Li}, \bibinfo{person}{M. Qiu},
  \bibinfo{person}{H. Chen}, \bibinfo{person}{X. Wang}, \bibinfo{person}{X.
  Gao}, \bibinfo{person}{J. Huang}, \bibinfo{person}{J. Ren},
  \bibinfo{person}{Z. Zhao}, \bibinfo{person}{W. Zhao}, \bibinfo{person}{L.
  Wang}, {and} \bibinfo{person}{G. Jin}.} \bibinfo{year}{2017}\natexlab{}.
\newblock \showarticletitle{AliMe Assist: An Intelligent Assistant for Creating
  an Innovative E-commerce Experience}. In \bibinfo{booktitle}{\emph{CIKM
  '17}}.
\newblock


\bibitem[\protect\citeauthoryear{Liu, Ott, Goyal, Du, Joshi, Chen, Levy, Lewis,
  Zettlemoyer, and Stoyanov}{Liu et~al\mbox{.}}{2019}]%
        {DBLP:journals/corr/abs-1907-11692}
\bibfield{author}{\bibinfo{person}{Yinhan Liu}, \bibinfo{person}{Myle Ott},
  \bibinfo{person}{Naman Goyal}, \bibinfo{person}{Jingfei Du},
  \bibinfo{person}{Mandar Joshi}, \bibinfo{person}{Danqi Chen},
  \bibinfo{person}{Omer Levy}, \bibinfo{person}{Mike Lewis},
  \bibinfo{person}{Luke Zettlemoyer}, {and} \bibinfo{person}{Veselin
  Stoyanov}.} \bibinfo{year}{2019}\natexlab{}.
\newblock \showarticletitle{RoBERTa: {A} Robustly Optimized {BERT} Pretraining
  Approach}.
\newblock \bibinfo{journal}{\emph{CoRR}}  \bibinfo{volume}{abs/1907.11692}
  (\bibinfo{year}{2019}).
\newblock
\showeprint[arxiv]{1907.11692}


\bibitem[\protect\citeauthoryear{Lv and Zhai}{Lv and Zhai}{2009}]%
        {Lv:2009:CSM:1645953.1646259}
\bibfield{author}{\bibinfo{person}{Y. Lv} {and} \bibinfo{person}{C. Zhai}.}
  \bibinfo{year}{2009}\natexlab{}.
\newblock \showarticletitle{A Comparative Study of Methods for Estimating Query
  Language Models with Pseudo Feedback}. In \bibinfo{booktitle}{\emph{CIKM
  '09}}.
\newblock


\bibitem[\protect\citeauthoryear{Mitra, Diaz, and Craswell}{Mitra
  et~al\mbox{.}}{2017}]%
        {Mitra:2017:LMU:3038912.3052579}
\bibfield{author}{\bibinfo{person}{B. Mitra}, \bibinfo{person}{F. Diaz}, {and}
  \bibinfo{person}{N. Craswell}.} \bibinfo{year}{2017}\natexlab{}.
\newblock \showarticletitle{Learning to Match Using Local and Distributed
  Representations of Text for Web Search}. In \bibinfo{booktitle}{\emph{WWW
  '17}}.
\newblock


\bibitem[\protect\citeauthoryear{Mnih, Kavukcuoglu, Silver, Rusu, Veness,
  Bellemare, Graves, Riedmiller, Fidjeland, Ostrovski, Petersen, Beattie,
  Sadik, Antonoglou, King, Kumaran, Wierstra, Legg, and Hassabis}{Mnih
  et~al\mbox{.}}{2015}]%
        {DBLP:journals/nature/MnihKSRVBGRFOPB15}
\bibfield{author}{\bibinfo{person}{Volodymyr Mnih}, \bibinfo{person}{Koray
  Kavukcuoglu}, \bibinfo{person}{David Silver}, \bibinfo{person}{Andrei~A.
  Rusu}, \bibinfo{person}{Joel Veness}, \bibinfo{person}{Marc~G. Bellemare},
  \bibinfo{person}{Alex Graves}, \bibinfo{person}{Martin~A. Riedmiller},
  \bibinfo{person}{Andreas Fidjeland}, \bibinfo{person}{Georg Ostrovski},
  \bibinfo{person}{Stig Petersen}, \bibinfo{person}{Charles Beattie},
  \bibinfo{person}{Amir Sadik}, \bibinfo{person}{Ioannis Antonoglou},
  \bibinfo{person}{Helen King}, \bibinfo{person}{Dharshan Kumaran},
  \bibinfo{person}{Daan Wierstra}, \bibinfo{person}{Shane Legg}, {and}
  \bibinfo{person}{Demis Hassabis}.} \bibinfo{year}{2015}\natexlab{}.
\newblock \showarticletitle{Human-level control through deep reinforcement
  learning}.
\newblock \bibinfo{journal}{\emph{Nature}} \bibinfo{volume}{518},
  \bibinfo{number}{7540} (\bibinfo{year}{2015}), \bibinfo{pages}{529--533}.
\newblock


\bibitem[\protect\citeauthoryear{Nogueira and Cho}{Nogueira and Cho}{[n.d.]}]%
        {passrerankbert19}
\bibfield{author}{\bibinfo{person}{Rodrigo Nogueira} {and}
  \bibinfo{person}{Kyunghyun Cho}.} \bibinfo{year}{[n.d.]}\natexlab{}.
\newblock \showarticletitle{Passage Re-ranking with {BERT}}.
\newblock \bibinfo{journal}{\emph{CoRR}} (\bibinfo{year}{[n.\,d.]}).
\newblock


\bibitem[\protect\citeauthoryear{Peters, Neumann, Iyyer, Gardner, Clark, Lee,
  and Zettlemoyer}{Peters et~al\mbox{.}}{2018}]%
        {Peters:2018}
\bibfield{author}{\bibinfo{person}{Matthew~E. Peters}, \bibinfo{person}{Mark
  Neumann}, \bibinfo{person}{Mohit Iyyer}, \bibinfo{person}{Matt Gardner},
  \bibinfo{person}{Christopher Clark}, \bibinfo{person}{Kenton Lee}, {and}
  \bibinfo{person}{Luke Zettlemoyer}.} \bibinfo{year}{2018}\natexlab{}.
\newblock \showarticletitle{Deep contextualized word representations}. In
  \bibinfo{booktitle}{\emph{Proc. of NAACL}}.
\newblock


\bibitem[\protect\citeauthoryear{Qiu, Li, Wang, Gao, Chen, Zhao, Chen, Huang,
  and Chu}{Qiu et~al\mbox{.}}{2017}]%
        {alime-chat}
\bibfield{author}{\bibinfo{person}{M. Qiu}, \bibinfo{person}{F. Li},
  \bibinfo{person}{S. Wang}, \bibinfo{person}{X. Gao}, \bibinfo{person}{Y.
  Chen}, \bibinfo{person}{W. Zhao}, \bibinfo{person}{H. Chen},
  \bibinfo{person}{J. Huang}, {and} \bibinfo{person}{W. Chu}.}
  \bibinfo{year}{2017}\natexlab{}.
\newblock \showarticletitle{AliMe Chat: A Sequence to Sequence and Rerank based
  Chatbot Engine}. In \bibinfo{booktitle}{\emph{ACL '17}}.
\newblock


\bibitem[\protect\citeauthoryear{Qiu, Li, Wang, Pan, Wang, Chen, Jia, Li,
  Huang, Cai, and Lin}{Qiu et~al\mbox{.}}{2021}]%
        {easytransfer}
\bibfield{author}{\bibinfo{person}{Minghui Qiu}, \bibinfo{person}{Peng Li},
  \bibinfo{person}{Chengyu Wang}, \bibinfo{person}{Haojie Pan},
  \bibinfo{person}{An Wang}, \bibinfo{person}{Cen Chen},
  \bibinfo{person}{Xianyan Jia}, \bibinfo{person}{Yaliang Li},
  \bibinfo{person}{Jun Huang}, \bibinfo{person}{Deng Cai}, {and}
  \bibinfo{person}{Wei Lin}.} \bibinfo{year}{2021}\natexlab{}.
\newblock \showarticletitle{EasyTransfer - A Simple and Scalable Deep Transfer
  Learning Platform for NLP Applications}.
\newblock \bibinfo{journal}{\emph{CIKM 2021}} (\bibinfo{year}{2021}).
\newblock
\urldef\tempurl%
\url{https://arxiv.org/abs/2011.09463}
\showURL{%
\tempurl}


\bibitem[\protect\citeauthoryear{Qu, Yang, Croft, Trippas, Zhang, and Qiu}{Qu
  et~al\mbox{.}}{2018}]%
        {DBLP:conf/sigir/QuYCTZQ18}
\bibfield{author}{\bibinfo{person}{C. Qu}, \bibinfo{person}{L. Yang},
  \bibinfo{person}{W.~B. Croft}, \bibinfo{person}{J.~R. Trippas},
  \bibinfo{person}{Y. Zhang}, {and} \bibinfo{person}{M. Qiu}.}
  \bibinfo{year}{2018}\natexlab{}.
\newblock \showarticletitle{Analyzing and Characterizing User Intent in
  Information-seeking Conversations}. In \bibinfo{booktitle}{\emph{SIGIR '18}}.
  \bibinfo{pages}{989--992}.
\newblock


\bibitem[\protect\citeauthoryear{Radford, Wu, Child, Luan, Amodei, and
  Sutskever}{Radford et~al\mbox{.}}{2019}]%
        {radford2019language}
\bibfield{author}{\bibinfo{person}{Alec Radford}, \bibinfo{person}{Jeff Wu},
  \bibinfo{person}{Rewon Child}, \bibinfo{person}{David Luan},
  \bibinfo{person}{Dario Amodei}, {and} \bibinfo{person}{Ilya Sutskever}.}
  \bibinfo{year}{2019}\natexlab{}.
\newblock \showarticletitle{Language Models are Unsupervised Multitask
  Learners}.
\newblock  (\bibinfo{year}{2019}).
\newblock


\bibitem[\protect\citeauthoryear{Radlinski and Craswell}{Radlinski and
  Craswell}{2017}]%
        {radlinski2017theoretical}
\bibfield{author}{\bibinfo{person}{F. Radlinski} {and} \bibinfo{person}{N.
  Craswell}.} \bibinfo{year}{2017}\natexlab{}.
\newblock \showarticletitle{A theoretical framework for conversational search}.
  In \bibinfo{booktitle}{\emph{CHIIR '17}}.
\newblock


\bibitem[\protect\citeauthoryear{Ritter, Cherry, and Dolan}{Ritter
  et~al\mbox{.}}{2011}]%
        {DBLP:conf/emnlp/RitterCD11}
\bibfield{author}{\bibinfo{person}{A. Ritter}, \bibinfo{person}{C. Cherry},
  {and} \bibinfo{person}{W.~B. Dolan}.} \bibinfo{year}{2011}\natexlab{}.
\newblock \showarticletitle{Data-Driven Response Generation in Social Media}.
  In \bibinfo{booktitle}{\emph{ACL '11}}.
\newblock


\bibitem[\protect\citeauthoryear{Robertson and Walker}{Robertson and
  Walker}{1994}]%
        {Robertson:1994:SEA:188490.188561}
\bibfield{author}{\bibinfo{person}{S. Robertson} {and} \bibinfo{person}{S.
  Walker}.} \bibinfo{year}{1994}\natexlab{}.
\newblock \showarticletitle{Some Simple Effective Approximations to the
  2-Poisson Model for Probabilistic Weighted Retrieval}. In
  \bibinfo{booktitle}{\emph{SIGIR '94}}.
\newblock


\bibitem[\protect\citeauthoryear{Rocchio}{Rocchio}{1971}]%
        {rocchio71relevance}
\bibfield{author}{\bibinfo{person}{J.~J. Rocchio}.}
  \bibinfo{year}{1971}\natexlab{}.
\newblock \showarticletitle{Relevance feedback in information retrieval}.
\newblock In \bibinfo{booktitle}{\emph{The Smart retrieval system - experiments
  in automatic document processing}},
  \bibfield{editor}{\bibinfo{person}{G.~Salton}} (Ed.).
\newblock


\bibitem[\protect\citeauthoryear{Rummery and Niranjan}{Rummery and
  Niranjan}{1994}]%
        {rummery:cuedtr94}
\bibfield{author}{\bibinfo{person}{G.~A. Rummery} {and} \bibinfo{person}{M.
  Niranjan}.} \bibinfo{year}{1994}\natexlab{}.
\newblock \bibinfo{booktitle}{\emph{On-Line {Q}-Learning Using Connectionist
  Systems}}.
\newblock \bibinfo{type}{{T}echnical {R}eport} TR 166.
  \bibinfo{institution}{Cambridge University Engineering Department},
  \bibinfo{address}{Cambridge, England}.
\newblock


\bibitem[\protect\citeauthoryear{Shang, Lu, and Li}{Shang
  et~al\mbox{.}}{2015}]%
        {DBLP:conf/acl/ShangLL15}
\bibfield{author}{\bibinfo{person}{L. Shang}, \bibinfo{person}{Z. Lu}, {and}
  \bibinfo{person}{H. Li}.} \bibinfo{year}{2015}\natexlab{}.
\newblock \showarticletitle{Neural Responding Machine for Short-Text
  Conversation}. In \bibinfo{booktitle}{\emph{ACL '15}}.
\newblock


\bibitem[\protect\citeauthoryear{Silver, Schrittwieser, Simonyan, Antonoglou,
  Huang, Guez, Hubert, Baker, Lai, Bolton, Chen, Lillicrap, Hui, Sifre, van~den
  Driessche, Graepel, and Hassabis}{Silver et~al\mbox{.}}{2017}]%
        {silver2017mastering}
\bibfield{author}{\bibinfo{person}{David Silver}, \bibinfo{person}{Julian
  Schrittwieser}, \bibinfo{person}{Karen Simonyan}, \bibinfo{person}{Ioannis
  Antonoglou}, \bibinfo{person}{Aja Huang}, \bibinfo{person}{Arthur Guez},
  \bibinfo{person}{Thomas Hubert}, \bibinfo{person}{Lucas Baker},
  \bibinfo{person}{Matthew Lai}, \bibinfo{person}{Adrian Bolton},
  \bibinfo{person}{Yutian Chen}, \bibinfo{person}{Timothy Lillicrap},
  \bibinfo{person}{Fan Hui}, \bibinfo{person}{Laurent Sifre},
  \bibinfo{person}{George van~den Driessche}, \bibinfo{person}{Thore Graepel},
  {and} \bibinfo{person}{Demis Hassabis}.} \bibinfo{year}{2017}\natexlab{}.
\newblock \showarticletitle{Mastering the game of Go without human knowledge}.
\newblock \bibinfo{journal}{\emph{Nature}}  \bibinfo{volume}{550}
  (\bibinfo{date}{Oct.} \bibinfo{year}{2017}), \bibinfo{pages}{354--}.
\newblock


\bibitem[\protect\citeauthoryear{Spina, Trippas, Cavedon, and Sanderson}{Spina
  et~al\mbox{.}}{2017}]%
        {spina2017extracting}
\bibfield{author}{\bibinfo{person}{D. Spina}, \bibinfo{person}{J.~R Trippas},
  \bibinfo{person}{L. Cavedon}, {and} \bibinfo{person}{M. Sanderson}.}
  \bibinfo{year}{2017}\natexlab{}.
\newblock \showarticletitle{Extracting audio summaries to support effective
  spoken document search}.
\newblock \bibinfo{journal}{\emph{JAIST '17}} \bibinfo{volume}{68},
  \bibinfo{number}{9} (\bibinfo{year}{2017}).
\newblock


\bibitem[\protect\citeauthoryear{Tao, Wu, Xu, Hu, Zhao, and Yan}{Tao
  et~al\mbox{.}}{2019}]%
        {Tao:2019:MFN:3289600.3290985}
\bibfield{author}{\bibinfo{person}{C. Tao}, \bibinfo{person}{W. Wu},
  \bibinfo{person}{C. Xu}, \bibinfo{person}{W. Hu}, \bibinfo{person}{D. Zhao},
  {and} \bibinfo{person}{R. Yan}.} \bibinfo{year}{2019}\natexlab{}.
\newblock \showarticletitle{Multi-Representation Fusion Network for Multi-Turn
  Response Selection in Retrieval-Based Chatbots}. In
  \bibinfo{booktitle}{\emph{WSDM '19}}.
\newblock


\bibitem[\protect\citeauthoryear{Thomas, McDu, Czerwinski, and Craswell}{Thomas
  et~al\mbox{.}}{2017}]%
        {thomas2017misc}
\bibfield{author}{\bibinfo{person}{P. Thomas}, \bibinfo{person}{D. McDu},
  \bibinfo{person}{M. Czerwinski}, {and} \bibinfo{person}{N. Craswell}.}
  \bibinfo{year}{2017}\natexlab{}.
\newblock \showarticletitle{MISC: A data set of information-seeking
  conversations}. In \bibinfo{booktitle}{\emph{CAIR '17}}.
\newblock


\bibitem[\protect\citeauthoryear{Trippas, Spina, Sanderson, and
  Cavedon}{Trippas et~al\mbox{.}}{2015}]%
        {trippas2015towards}
\bibfield{author}{\bibinfo{person}{J. Trippas}, \bibinfo{person}{D. Spina},
  \bibinfo{person}{M. Sanderson}, {and} \bibinfo{person}{L. Cavedon}.}
  \bibinfo{year}{2015}\natexlab{}.
\newblock \showarticletitle{Towards understanding the impact of length in web
  search result summaries over a speech-only communication channel}. In
  \bibinfo{booktitle}{\emph{SIGIR '15}}.
\newblock


\bibitem[\protect\citeauthoryear{Vig and Ramea}{Vig and Ramea}{2019}]%
        {JesseVig19}
\bibfield{author}{\bibinfo{person}{Jesse Vig} {and} \bibinfo{person}{Kalai
  Ramea}.} \bibinfo{year}{2019}\natexlab{}.
\newblock \showarticletitle{Comparison of Transfer-Learning Approaches for
  Response Selection in Multi-Turn Conversations}.
\newblock  (\bibinfo{year}{2019}).
\newblock


\bibitem[\protect\citeauthoryear{Vinyals and Le}{Vinyals and Le}{2015}]%
        {DBLP:journals/corr/VinyalsL15}
\bibfield{author}{\bibinfo{person}{O. Vinyals} {and} \bibinfo{person}{Q.~V.
  Le}.} \bibinfo{year}{2015}\natexlab{}.
\newblock \showarticletitle{A Neural Conversational Model}.
\newblock \bibinfo{journal}{\emph{CoRR}}  \bibinfo{volume}{abs/1506.05869}
  (\bibinfo{year}{2015}).
\newblock


\bibitem[\protect\citeauthoryear{Wan, Lan, Guo, Xu, Pang, and Cheng}{Wan
  et~al\mbox{.}}{2016}]%
        {DBLP:conf/aaai/WanLGXPC16}
\bibfield{author}{\bibinfo{person}{S. Wan}, \bibinfo{person}{Y. Lan},
  \bibinfo{person}{J. Guo}, \bibinfo{person}{J. Xu}, \bibinfo{person}{L. Pang},
  {and} \bibinfo{person}{X. Cheng}.} \bibinfo{year}{2016}\natexlab{}.
\newblock \showarticletitle{A Deep Architecture for Semantic Matching with
  Multiple Positional Sentence Representations}. In
  \bibinfo{booktitle}{\emph{AAAI '16}}.
\newblock


\bibitem[\protect\citeauthoryear{Wang, Yu, Guo, Wang, Klinger, Zhang, Chang,
  Tesauro, Zhou, and Jiang}{Wang et~al\mbox{.}}{[n.d.]}]%
        {DBLP:conf/aaai/WangYGWKZCTZJ18}
\bibfield{author}{\bibinfo{person}{Shuohang Wang}, \bibinfo{person}{Mo Yu},
  \bibinfo{person}{Xiaoxiao Guo}, \bibinfo{person}{Zhiguo Wang},
  \bibinfo{person}{Tim Klinger}, \bibinfo{person}{Wei Zhang},
  \bibinfo{person}{Shiyu Chang}, \bibinfo{person}{Gerry Tesauro},
  \bibinfo{person}{Bowen Zhou}, {and} \bibinfo{person}{Jing Jiang}.}
  \bibinfo{year}{[n.d.]}\natexlab{}.
\newblock \showarticletitle{R\({}^{\mbox{3}}\): Reinforced Ranker-Reader for
  Open-Domain Question Answering}. In \bibinfo{booktitle}{\emph{AAAI}}.
  \bibinfo{pages}{5981--5988}.
\newblock


\bibitem[\protect\citeauthoryear{Whang, Lee, Lee, Yang, Oh, and Lim}{Whang
  et~al\mbox{.}}{2019}]%
        {DBLP:journals/corr/abs-1908-04812}
\bibfield{author}{\bibinfo{person}{Taesun Whang}, \bibinfo{person}{Dongyub
  Lee}, \bibinfo{person}{Chanhee Lee}, \bibinfo{person}{Kisu Yang},
  \bibinfo{person}{Dongsuk Oh}, {and} \bibinfo{person}{Heuiseok Lim}.}
  \bibinfo{year}{2019}\natexlab{}.
\newblock \showarticletitle{Domain Adaptive Training {BERT} for Response
  Selection}.
\newblock \bibinfo{journal}{\emph{CoRR}} (\bibinfo{year}{2019}).
\newblock


\bibitem[\protect\citeauthoryear{Williams}{Williams}{1992}]%
        {DBLP:journals/ml/Williams92}
\bibfield{author}{\bibinfo{person}{Ronald~J. Williams}.}
  \bibinfo{year}{1992}\natexlab{}.
\newblock \showarticletitle{Simple Statistical Gradient-Following Algorithms
  for Connectionist Reinforcement Learning}.
\newblock \bibinfo{journal}{\emph{Mach. Learn.}}  \bibinfo{volume}{8}
  (\bibinfo{year}{1992}), \bibinfo{pages}{229--256}.
\newblock


\bibitem[\protect\citeauthoryear{Wu, Li, and Wang}{Wu et~al\mbox{.}}{[n.d.]}]%
        {DBLP:conf/naacl/WuLW18}
\bibfield{author}{\bibinfo{person}{Jiawei Wu}, \bibinfo{person}{Lei Li}, {and}
  \bibinfo{person}{William~Yang Wang}.} \bibinfo{year}{[n.d.]}\natexlab{}.
\newblock \showarticletitle{Reinforced Co-Training}. In
  \bibinfo{booktitle}{\emph{NAACL-HLT}}. \bibinfo{pages}{1252--1262}.
\newblock


\bibitem[\protect\citeauthoryear{Wu, Schuster, Chen, Le, Norouzi, Macherey,
  Krikun, Cao, Gao, Macherey, Klingner, Shah, Johnson, Liu, Kaiser, Gouws, and
  et~al.}{Wu et~al\mbox{.}}{2016}]%
        {DBLP:journals/corr/WuSCLNMKCGMKSJL16}
\bibfield{author}{\bibinfo{person}{Yonghui Wu}, \bibinfo{person}{Mike
  Schuster}, \bibinfo{person}{Zhifeng Chen}, \bibinfo{person}{Quoc~V. Le},
  \bibinfo{person}{Mohammad Norouzi}, \bibinfo{person}{Wolfgang Macherey},
  \bibinfo{person}{Maxim Krikun}, \bibinfo{person}{Yuan Cao},
  \bibinfo{person}{Qin Gao}, \bibinfo{person}{Klaus Macherey},
  \bibinfo{person}{Jeff Klingner}, \bibinfo{person}{Apurva Shah},
  \bibinfo{person}{Melvin Johnson}, \bibinfo{person}{Xiaobing Liu},
  \bibinfo{person}{Lukasz Kaiser}, \bibinfo{person}{Stephan Gouws}, {and}
  \bibinfo{person}{Jeffrey~Dean et al.}} \bibinfo{year}{2016}\natexlab{}.
\newblock \showarticletitle{Google's Neural Machine Translation System:
  Bridging the Gap between Human and Machine Translation}.
\newblock \bibinfo{journal}{\emph{CoRR}} (\bibinfo{year}{2016}).
\newblock


\bibitem[\protect\citeauthoryear{Wu, Wu, Xing, Zhou, and Li}{Wu
  et~al\mbox{.}}{2017}]%
        {DBLP:conf/acl/WuWXZL17}
\bibfield{author}{\bibinfo{person}{Y. Wu}, \bibinfo{person}{W. Wu},
  \bibinfo{person}{C. Xing}, \bibinfo{person}{M. Zhou}, {and}
  \bibinfo{person}{Z. Li}.} \bibinfo{year}{2017}\natexlab{}.
\newblock \showarticletitle{Sequential Matching Network: {A} New Architecture
  for Multi-turn Response Selection in Retrieval-Based Chatbots}. In
  \bibinfo{booktitle}{\emph{ACL '17}}.
\newblock


\bibitem[\protect\citeauthoryear{Xu, Liu, Shu, and Yu}{Xu
  et~al\mbox{.}}{[n.d.]}]%
        {DBLP:conf/naacl/XuLSY19}
\bibfield{author}{\bibinfo{person}{Hu Xu}, \bibinfo{person}{Bing Liu},
  \bibinfo{person}{Lei Shu}, {and} \bibinfo{person}{Philip~S. Yu}.}
  \bibinfo{year}{[n.d.]}\natexlab{}.
\newblock \showarticletitle{{BERT} Post-Training for Review Reading
  Comprehension and Aspect-based Sentiment Analysis}. In
  \bibinfo{booktitle}{\emph{NAACL-HLT}}. \bibinfo{pages}{2324--2335}.
\newblock


\bibitem[\protect\citeauthoryear{Yan, Song, and Wu}{Yan et~al\mbox{.}}{2016}]%
        {DBLP:conf/sigir/YanSW16}
\bibfield{author}{\bibinfo{person}{R. Yan}, \bibinfo{person}{Y. Song}, {and}
  \bibinfo{person}{H. Wu}.} \bibinfo{year}{2016}\natexlab{}.
\newblock \showarticletitle{Learning to Respond with Deep Neural Networks for
  Retrieval-Based Human-Computer Conversation System}. In
  \bibinfo{booktitle}{\emph{SIGIR}}.
\newblock


\bibitem[\protect\citeauthoryear{Yan, Zhao, and E.}{Yan et~al\mbox{.}}{2017}]%
        {DBLP:conf/sigir/YanZE17}
\bibfield{author}{\bibinfo{person}{R. Yan}, \bibinfo{person}{D. Zhao}, {and}
  \bibinfo{person}{W. E.}} \bibinfo{year}{2017}\natexlab{}.
\newblock \showarticletitle{Joint Learning of Response Ranking and Next
  Utterance Suggestion in Human-Computer Conversation System}. In
  \bibinfo{booktitle}{\emph{SIGIR '17}}.
\newblock


\bibitem[\protect\citeauthoryear{Yang, Qiu, Qu, Chen, Guo, Zhang, Croft, and
  Chen}{Yang et~al\mbox{.}}{2020}]%
        {DBLP:conf/www/0005QQCGZCC20}
\bibfield{author}{\bibinfo{person}{Liu Yang}, \bibinfo{person}{Minghui Qiu},
  \bibinfo{person}{Chen Qu}, \bibinfo{person}{Cen Chen},
  \bibinfo{person}{Jiafeng Guo}, \bibinfo{person}{Yongfeng Zhang},
  \bibinfo{person}{W.~Bruce Croft}, {and} \bibinfo{person}{Haiqing Chen}.}
  \bibinfo{year}{2020}\natexlab{}.
\newblock \showarticletitle{{IART:} Intent-aware Response Ranking with
  Transformers in Information-seeking Conversation Systems}. In
  \bibinfo{booktitle}{\emph{WWW}}. \bibinfo{pages}{2592--2598}.
\newblock


\bibitem[\protect\citeauthoryear{Yang, Qiu, Qu, Guo, Zhang, Croft, Huang, and
  Chen}{Yang et~al\mbox{.}}{2018}]%
        {DBLP:conf/sigir/YangQQGZCHC18}
\bibfield{author}{\bibinfo{person}{L. Yang}, \bibinfo{person}{M. Qiu},
  \bibinfo{person}{C. Qu}, \bibinfo{person}{J. Guo}, \bibinfo{person}{Y.
  Zhang}, \bibinfo{person}{W.~B. Croft}, \bibinfo{person}{J. Huang}, {and}
  \bibinfo{person}{H. Chen}.} \bibinfo{year}{2018}\natexlab{}.
\newblock \showarticletitle{Response Ranking with Deep Matching Networks and
  External Knowledge in Information-seeking Conversation Systems}. In
  \bibinfo{booktitle}{\emph{SIGIR '18}}.
\newblock


\bibitem[\protect\citeauthoryear{Yang, Zhang, and Lin}{Yang
  et~al\mbox{.}}{2019b}]%
        {SimpleAppBERTDocRetreive19}
\bibfield{author}{\bibinfo{person}{Wei Yang}, \bibinfo{person}{Haotian Zhang},
  {and} \bibinfo{person}{Jimmy Lin}.} \bibinfo{year}{2019}\natexlab{b}.
\newblock \showarticletitle{Simple Applications of {BERT} for Ad Hoc Document
  Retrieval}.
\newblock \bibinfo{journal}{\emph{CoRR}} (\bibinfo{year}{2019}).
\newblock


\bibitem[\protect\citeauthoryear{Yang, Dai, Yang, Carbonell, Salakhutdinov, and
  Le}{Yang et~al\mbox{.}}{2019a}]%
        {DBLP:conf/nips/YangDYCSL19}
\bibfield{author}{\bibinfo{person}{Zhilin Yang}, \bibinfo{person}{Zihang Dai},
  \bibinfo{person}{Yiming Yang}, \bibinfo{person}{Jaime~G. Carbonell},
  \bibinfo{person}{Ruslan Salakhutdinov}, {and} \bibinfo{person}{Quoc~V. Le}.}
  \bibinfo{year}{2019}\natexlab{a}.
\newblock \showarticletitle{XLNet: Generalized Autoregressive Pretraining for
  Language Understanding}. In \bibinfo{booktitle}{\emph{NeurIPS}}.
  \bibinfo{pages}{5754--5764}.
\newblock


\bibitem[\protect\citeauthoryear{Zamani, Dadashkarimi, Shakery, and
  Croft}{Zamani et~al\mbox{.}}{2016}]%
        {Zamani:2016:PFB:2983323.2983844}
\bibfield{author}{\bibinfo{person}{H. Zamani}, \bibinfo{person}{J.
  Dadashkarimi}, \bibinfo{person}{A. Shakery}, {and} \bibinfo{person}{W.~B.
  Croft}.} \bibinfo{year}{2016}\natexlab{}.
\newblock \showarticletitle{Pseudo-Relevance Feedback Based on Matrix
  Factorization}. In \bibinfo{booktitle}{\emph{CIKM '16}}.
\newblock


\bibitem[\protect\citeauthoryear{Zhai and Lafferty}{Zhai and Lafferty}{2001}]%
        {Zhai:2001:MFL:502585.502654}
\bibfield{author}{\bibinfo{person}{C. Zhai} {and} \bibinfo{person}{J.
  Lafferty}.} \bibinfo{year}{2001}\natexlab{}.
\newblock \showarticletitle{Model-based Feedback in the Language Modeling
  Approach to Information Retrieval}. In \bibinfo{booktitle}{\emph{CIKM '01}}.
\newblock


\bibitem[\protect\citeauthoryear{Zhang, Chen, Ai, Yang, and Croft}{Zhang
  et~al\mbox{.}}{2018}]%
        {zhang2018towards}
\bibfield{author}{\bibinfo{person}{Y. Zhang}, \bibinfo{person}{X. Chen},
  \bibinfo{person}{Q. Ai}, \bibinfo{person}{L. Yang}, {and}
  \bibinfo{person}{W.~B. Croft}.} \bibinfo{year}{2018}\natexlab{}.
\newblock \showarticletitle{Towards Conversational Search and Recommendation:
  System Ask, User Respond}. In \bibinfo{booktitle}{\emph{CIKM '18}}.
\newblock


\bibitem[\protect\citeauthoryear{Zhou, Dong, Wu, Zhao, Yu, Tian, Liu, and
  Yan}{Zhou et~al\mbox{.}}{2016}]%
        {DBLP:conf/emnlp/ZhouDWZYTLY16}
\bibfield{author}{\bibinfo{person}{X. Zhou}, \bibinfo{person}{D. Dong},
  \bibinfo{person}{H. Wu}, \bibinfo{person}{S. Zhao}, \bibinfo{person}{D. Yu},
  \bibinfo{person}{H. Tian}, \bibinfo{person}{X. Liu}, {and}
  \bibinfo{person}{R. Yan}.} \bibinfo{year}{2016}\natexlab{}.
\newblock \showarticletitle{Multi-view Response Selection for Human-Computer
  Conversation}. In \bibinfo{booktitle}{\emph{EMNLP}}.
\newblock


\bibitem[\protect\citeauthoryear{Zhou, Li, Dong, Liu, Chen, Zhao, Yu, and
  Wu}{Zhou et~al\mbox{.}}{2018}]%
        {DBLP:conf/acl/WuLCZDYZL18}
\bibfield{author}{\bibinfo{person}{X. Zhou}, \bibinfo{person}{L. Li},
  \bibinfo{person}{D. Dong}, \bibinfo{person}{Y. Liu}, \bibinfo{person}{Y.
  Chen}, \bibinfo{person}{W.~X. Zhao}, \bibinfo{person}{D. Yu}, {and}
  \bibinfo{person}{H. Wu}.} \bibinfo{year}{2018}\natexlab{}.
\newblock \showarticletitle{Multi-Turn Response Selection for Chatbots with
  Deep Attention Matching Network}. In \bibinfo{booktitle}{\emph{ACL '18}}.
\newblock


\end{thebibliography}


\end{document}